# A Macrocolumn Architecture Implemented with Spiking Neurons

J. E. Smith

University of Wisconsin-Madison (Emeritus)

Feb. 09, 2023

## Abstract

The *macrocolumn* is a key component of a neuromorphic computing system that interacts with an external *environment* under control of an *agent*. Environments are learned and stored in the macrocolumn as labeled directed graphs where edges connect features and labels indicate the relative displacements between them.

Macrocolumn functionality is first defined with a state machine model. This model is then implemented with a neural network composed of spiking neurons. The neuron model employs active dendrites and mirrors the Hawkins/Numenta neuron model. The architecture is demonstrated with a research benchmark in which an agent employs a macrocolumn to first learn and then navigate 2-d environments containing pseudo-randomly placed features.

## 1. Introduction

The neocortex is an essential part of any biologically plausible brain model. The biological neocortex is constructed of a few million macrocolumns, each consisting of about 100 minicolumns. As put forward by Mountcastle [20], the basic macrocolumn structure is uniform across the neocortex. Although specific connectivities and neuron characteristics may differ across cortical regions, macrocolumn operating principles remain the same.

In this paper, macrocolumn functions and implementations are studied in the context of an overall system containing an agent that interacts with a model environment. The role of the agent is described, although only a very primitive agent is actually implemented in the research benchmark; the focus here is on the macrocolumn.

A promising approach to macrocolumn architecture is advocated by Jeff Hawkins and his colleagues at Numenta [8]-[12][14], and this paper marks a transition in the author's research toward the Hawkins/Numenta approach to neuron operation and macrocolumn design. By incorporating the concept of *active dendrites* [9], the "spiking neuron" in prior work [25][26][27] is relegated to a spiking dendritic segment in this document. A dendrite is then composed of multiple segments, and a neuron is composed of multiple dendrites. As part of the active dendrite model, neuron inputs are categorized as being *distal* and *proximal*, each playing a role in macrocolumn operation.

Hawkins [12] accepts the Mountcastle uniformity hypothesis and focuses on four principles of functional uniformity:

      1) *Reference frames are present everywhere in the neocortex.*
      2) *Reference frames are used to model everything we know, not just physical objects.*
      3) *All knowledge is stored at locations relative to reference frames.*
      4) *Thinking is a form of movement.*

The ubiquitous reference frames are implemented by macrocolumns. The Mountcastle uniformity hypothesis and Hawkins's four principles underpin the modeling approach in this document.

Hodgkin and Huxley [13] developed their famous neuron model by studying squid giant axons because their size makes them good candidates for physical experimentation. Although their research focused on a particular type of easy-to-study neuron, the fundamental principles apply to all neurons. Following that philosophy, in this document a basic cognitive architecture is constructed and applied to an easy-to-understand research benchmark that involves movement and feature learning in a two dimensional physical space. For this application, the reference frame is a two dimensional grid, and knowledge is the presence of features at relative grid locations. From a researcher's perspective, a 2-d grid application is easy to understand, explain, and simulate.

"Thinking is a form of movement" applies to all forms of thinking including higher level thinking, not just navigation of physical environments (see [11], p. 11). As two examples, "thinking" includes mathematical reasoning and language processing. Taken together, the uniformity hypothesis and "thinking is a form of movement" strongly suggest that the same structure and principles displayed in the 2-d grid-based application developed in this paper may apply, at some fundamental level, to all forms of thinking.

## 1.1 System Architecture

Macrocolumns are developed and studied as part of an overall system organization depicted in Figure 1. An *agent* generates actions that cause *movement* in an *environment*. As movement occurs, a macrocolumn consisting of grid and place cells associates sensed environmental *features* with relative grid locations.

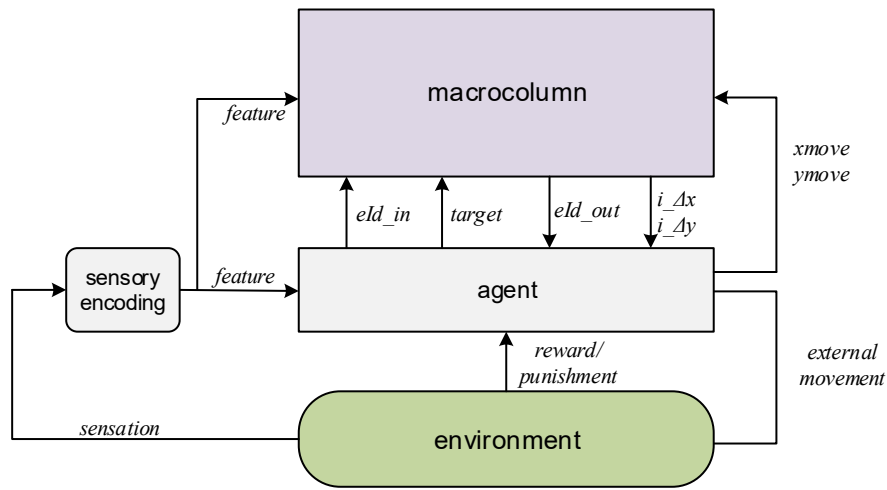

**Figure 1. The basic system architecture contains a *macrocolumn* that supports an *agent* in pursuit of some objective. The agent issues movement commands as it navigates a grid-based physical *environment*. For a given environment identifier (*eId in*), the agent learns an association between an externally sensed *feature*, the previous feature, and the relative displacement ($\Delta x$ and $\Delta y$) between the two. The previous feature and displacement are maintained by a grid subsystem within the macrocolumn. Then, when queried by the agent (by providing a *target feature*) the macrocolumn returns the *inferred displacement* $i\_\Delta x$ and $i\_\Delta y$ between the current *feature* and the *target feature*. This displacement allows the agent to move directly from its current location to a *target feature*'s location via both *internal* and *external movement* signals. Internal movement, *xmove* and *ymove*, affects the internal grid subsystem, and *external movement* affects the location in the physical environment. *Internal movements* track *external movements* by design.**

The model agent invokes both *internal* and *external movements*. If the agent is directing movement through a physical environment, then an external movement is tracked by an isomorphic movement along the internal grid. However, in some instances the agent may invoke an internal movement only. This might be done when performing "mind travel" [23], for example, when the agent wants to be aware of features within its current environment without actually performing the physical movements. Based on the outcome of such mind travel the agent may or may not proceed with the actual external movement.

Finally, as noted above, the same basic system architecture can be applied to non-physical environments. In that case, all movements may be internal movements.

In the system architecture proposed here, the macrocolumn does not make movement decisions, rather the agent does, so the macrocolumn plays a role loosely analogous to a Q-Table in reinforcement learning [33]. As the agent directs movement through an environment, the macrocolumn dynamically updates synaptic weights that collectively characterize environmental features and their relative locations as they are encountered. After adequate learning, the synaptic weights, through an inference process, provide information to the agent as it tries to accomplish a specific objective, e.g., *inferred displacements* ($i\_\Delta x$ and $i\_\Delta y$) and *environment identifiers* (*eId_out*) in the figure.

### *1.2 Research Benchmark*

Macrocolumn architecture is demonstrated by constructing a system as shown in Figure 1 containing a single macrocolumn. A demonstration system capable of executing a benchmark suggested by Lewis et al. [14] is designed and verified via simulation.

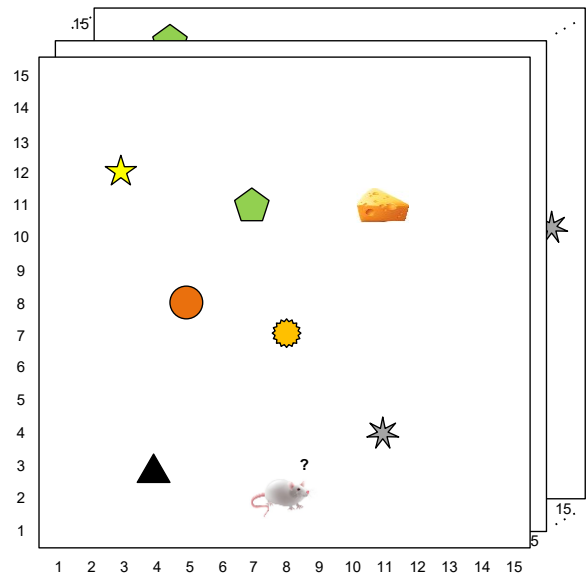

**Figure 2. "Mouse in the Dark" A mouse first learns multiple environments, each containing randomly placed features. After learning, when placed at a random place in a random environment, the mouse orients itself by identifying its environment and then uses learned features to navigate the environment, for example to find a specific feature such as a piece of cheese.**

The research benchmark involves a mouse navigating 2-dimensional environments in the dark (only features at the current location can be sensed). Because a single macrocolumn can hold multiple environments at the same time, the mouse is introduced to a sequence of environments, say a few tens, and has the opportunity to explore and learn each of them. As initial explorations take place, the macrocolumn learns spatial relationships among features within each of the environments.

Then, the mouse is dropped into an arbitrary environment at an arbitrary location. It first orients itself by moving about in the environment, associating features it senses with what it previously learned. Assuming no inherent ambiguity in the environments, this process eventually converges to a unique feature within a known environment: the mouse is oriented.

In this paper, a naive agent is used to illustrate the basic ability to acquire navigational information from the macrocolumn. Although not described here, a more elaborate agent employing reinforcement learning is under development. For example, one of the features may be "cheese", so if the mouse is placed in a learned environment in a "hungry" state, the agent employs a navigational strategy that uses the macrocolumn to first orient itself and then find the cheese.

To summarize, the macrocolumn supports three basic tasks under the direction of an agent:

    1) Exploration: It learns environments by encountering features while exploring.

    2) Orientation: When placed in an arbitrary learned environment, it orients itself.

    3) Queries to support navigation: After orientation has taken place, the agent can query the macrocolumn to acquire navigational information.

This paper constructs a macrocolumn architecture composed of biologically plausible spiking neurons that provides these three functions in a straightforward manner.

### *1.3  Contributions*

The main contributions are the following.

1) A state machine description of macrocolumn operation – The state machine constructs a directed graph representation of environments that is stored in an associative memory.

2) The implementation of model neurons based on active dendrites – Dendritic segments take binary (spike) inputs and integrate synaptic weights, but rather than generating and outputting a spike when a threshold is reached, as is commonly done, the total weight is output instead.

3) Construction and demonstration of a macrocolumn that implements the state machine description with model neurons – This is done bottom up by developing models for dendritic segments, dendrites, neurons, minicolumns, and macrocolumns. It is proposed that the macrocolumn uses loops formed by autaptic connections as a non-synaptic way of maintaining state for extended periods of time.

## 2. State Machine Description

In this section, a state machine definition of macrocolumn operation is developed. This functional definition is easy to understand and in following sections serves as a specification for a detailed macrocolumn implementation using neural networks. First the research benchmark problem is described. Then, the state machine description is given in some detail, using the research benchmark as an extended illustrative example.

### *2.1 Research Benchmark Specifics*

An environment, labeled "α" is shown in Figure 3a. A mouse (controlled by the agent) is dropped into the environment at an arbitrary location ("start" in Figure 3b) and begins exploring the environment at the agent's direction. The agent explores randomly, encountering features along the way. As it does so, it keeps track of its movements by maintaining the net displacements ($\Delta x$ and $\Delta y$) from the most recent feature it has encountered. When it reaches the next feature, it uses this information to construct and learn an edge in a directed graph representation of the environment (Figure 3c). During exploration it may encounter the same feature more than once, so a given feature in the graph may have in-degree greater than one. After an adequate amount of learning, the mouse is introduced to a second environment β which it also learns. Here, "adequate amount of learning" is loosely defined. At a minimum it should learn a connected graph, so every feature is reachable from every other feature. On the other hand, a complete graph is likely to be too time-consuming to learn and too expensive to implement except for the simplest environments.

This process is repeated for a second environment β shown in Figure 4. In Section 8, a total of 40 environments, each 30×30, are simulated.

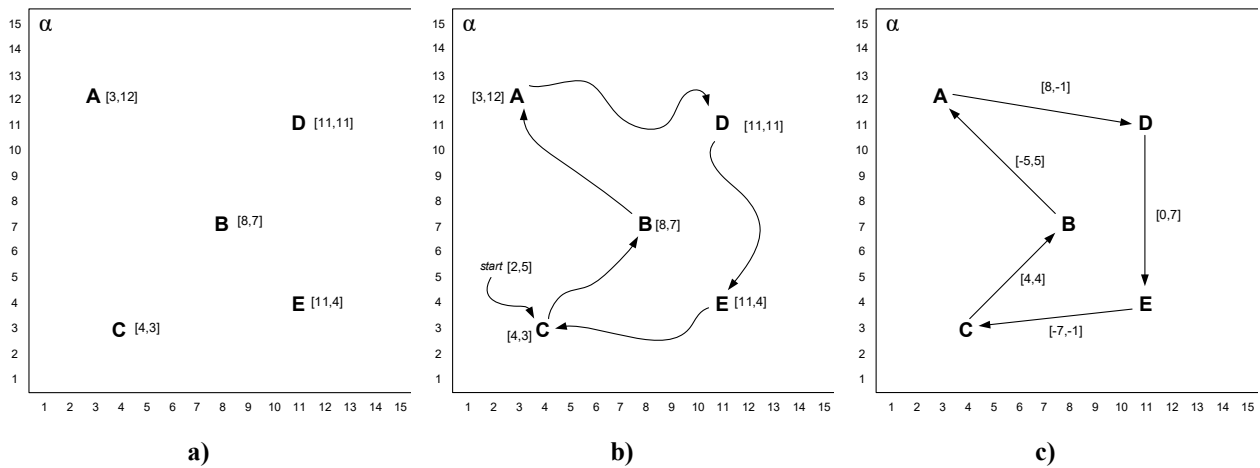

**Figure 3. a) The environment α encompasses a 15x15 grid containing randomly placed features A through E. The coordinates of each feature are shown. b) Beginning at an arbitrary starting point, the mouse explores the environment. c) As it explores, it learns the features and their relative locations by constructing a directed graph labeled with displacements from one feature to another. In general, the same feature can be visited more than once.**

J. E. Smith 5 02/09/2023

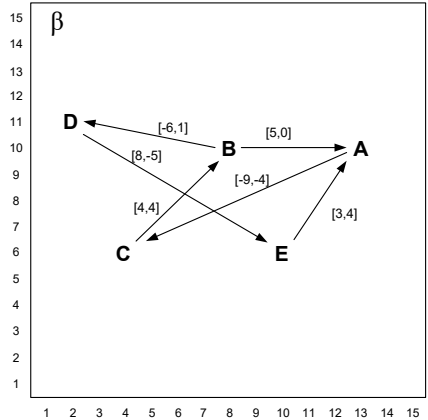

**Figure 4. A second environment, β, with learned edges.**

After the initial exploration/learning process is complete, the mouse is dropped into an arbitrary environment at an arbitrary location. It orients itself by moving about the environment, associating movements and features it senses with the previously learned graphs. As long as the environments are theoretically resolvable using the learned information, this process eventually converges to a unique feature within a known environment, and the mouse is oriented.

After orientation, the agent can use the learned information to navigate the environment. If the agent is located at a given feature and applies a desired destination (i.e., a target feature) to the macrocolumn, the macrocolumn produces the displacement to the destination feature as an output – provided the macrocolumn previously learned the labeled edge between the two features. This displacement information allows the agent to move directly to the desired destination feature.

### *2.2 State Machine Implementation*

The state machine implementation maintains a state vector that is constantly updated as the agent directs movement through an environment. Specifically, the state machine consists of 1) a state vector, 2) state vector update functions, and 3) an associative memory that stores learned state vectors. The implementation has three operating modes:

*Exploration*: the agent moves the state machine through the environment, and the state machine constructs (learns) the directed graph environment as it goes. The directed graph is stored as a set of state vectors held in the associative memory.

*Orientation*: beginning at an unknown environment and location, the agent directs movement through the environment, and accesses the associative memory as part of a process that determines its environment and location.

*Queries* (*Navigation*): the agent queries the associative memory with a target feature, and the associative memory returns the displacement from the current feature to the target feature (if it has been learned).

In discussions to follow, the modes are described and illustrated separately. However, in practice, they are not mutually exclusive. For example, exploration (learning) can potentially occur concurrently with navigation.

*State Vector*

As the agent moves through an environment, a *state vector* consisting of five components is maintained within a macrocolumn. The state vector contains all information necessary for constructing labeled directed edges in an environment graph. The components are:

    *eId:* the environment identifier
    *tail* : the feature at the tail of the directed edge
    $\Delta x$ : the *x* dimension displacement between *tail* and *head*
    $\Delta y$: the *y* dimension displacement between *tail* and *head*
    *head* : the feature at the head of the directed edge

A state vector is written as: [ *eId  tail  $\Delta x$  $\Delta y$ head* ]. An example state vector taken from environment α in Figure 3 is: [α C 4 4 B]. At any given time some of the vector components may be *null*, for example the two displacements are *null* in state vector [α C - - B]. If the agent is at a location that does not contain a feature then *head* is *null*. Furthermore, some components may be ambiguous; for example *eId* is ambiguous in [α,β C 4 4 B].

*State Machine Inputs*

    *eId_in* – the current environment from the agent when in exploration mode; otherwise null
    *target* – from the agent when making a query
    *xmove* – *x* dimension movement, from the agent
    *ymove* – *y* dimension movement, from the agent
    *feature* – environmental feature at the current grid location (if there is a feature, otherwise *null*)

*State Machine Outputs*

    *eId_out* – to the agent; *eId_out* is set to the inferred *i_eId* whenever *i_eId* ~= *null*.
    *i_$\Delta x$* – inferred $\Delta x$, to the agent
    *i_$\Delta y$* – inferred $\Delta y$, to the agent

*Associative Memory*

Because the associative memory is implemented as *place* cells in the macrocolumn to be described later, it is denoted here as *PL*. As the agent explores the environment to learn the graph representation, the associative memory *PL* stores edges in the form of state vectors:

    *PL* ← [*state vector*].

This is essentially a form of one-shot learning – it is simply a write to the associative memory. However, it is important to note that in the *PL* implementation in Section 7, learning is a biologically plausible, incremental update process that accounts for both the current state and what has been learned from previous states. This is the only significant difference between the state machine description given here and the spiking neuron macrocolumn implementation.

Every cycle, *PL* produces three inferred outputs: *i_eId*, *i_$\Delta x$,* and *i_$\Delta y$*, although they are not actually used every cycle. The *i_eId* output is used during the orientation process, and the inferred displacements are used during the query/navigation process.

    [*i_eId, i_$\Delta x$, i_$\Delta y$*] ← *PL* ([*state vector*])

The outputs are all *null* if the head component of the state vector is *null*. Otherwise, it produces outputs from the *PL* entry whose components most closely match the *state vector's* components.

*State Update Functions*

As the state machine moves through the environment under the direction of the agent, it may encounter a *feature* at a given grid location. The *feature* serves as the state vector *head*. If there is no feature at a given location, then *feature = null*.

At each time step, the state vector is updated. State vector update functions for each of the components of the state vector follow. Technically, the variable *eId_out* is state that is carried from one cycle to the next.

    *eId_out*: if $i\_eId \mathrel{\sim}= null$   $eId\_out \leftarrow i\_eId$
          else   $eId\_out \leftarrow eId\_out$

The actual *eId* that is part of the state vector fed to PL depends on the mode:
    *eId*:    if $explore == 1$   $eId \leftarrow eId\_in$
          else $eId \leftarrow eId\_out$

The *tail* is essentially maintained in its own small state machine:
    *tail*:    if $feature \mathrel{\sim}= null$   $tail \leftarrow feature$
          else $tail \leftarrow tail$

The displacements are reset whenever a new feature is encountered and then accumulate movements until the next feature is reached.
    $\Delta x, \Delta y$: if $feature \mathrel{\sim}= null$, $\Delta x \leftarrow null$, $\Delta y \leftarrow null$   (*reset*)
          else $\Delta x \leftarrow \Delta x + xmove$; $\Delta y \leftarrow \Delta y + ymove$

The *head* is not state that is maintained internally as the other state variables are, rather it depends on the *feature* at the current grid location or is set by the agent to some *target* feature as part of a query.
    *head*:  if $query == 1$   $head \leftarrow target$
          else   $head \leftarrow feature$

## 2.3 Detailed Example

*Exploration*

Figure 5 illustrates a sequence of movements during the exploration of environment α. Initially, all state vector components except *eId* are *null*. (In state vectors, *null* is represented as "-".) The agent provides *eId_in,* the identifier for the environment about to be learned. The *eId_in* can be any arbitrary coding; the only thing that is important is that each environment has its own unique Id.

As the agent moves through the grid, the state vector is unchanged until the first feature, C, is reached. Whenever the current grid location contains a feature, the displacements are initialized to $\Delta x, \Delta y = $ [-,-], so the state vector becomes [α - - - C]. Next, there is a single time step pause (zero movement). At the beginning of the pause step, *head* $\mathrel{\sim}= null$, so *tail* $\leftarrow$ *head*, and at the end of the step, feature C is at both the *head* and the *tail,* and the relative displacements are *null*.

Then, the agent moves away from the grid location containing C onto a location with no features, so *head* $\leftarrow$ *null* and *tail* $\leftarrow$ *tail*. The movement [1,1] yields state vector [α C 1 1 -]. After some additional moves, feature B is reached. In the example sequence, this is shown as a single move [3,3]. At that point the state vector components are all non-*null*: [α C 4 4 B], and the state vector is a complete description of the edge connecting C and B in environment α. The presence of a complete state vector triggers its storage in the associative memory *PL*.

Exploration continues in the same manner, and *PL* stores each new edge as it is discovered. Similarly, the agent explores environment β. After exploration, the contents of *PL* are shown in Figure 6.

*State Vector Sequence: Exploration*

| movement | state vector | comments |
|---|---|---|
| – – | α – – – – | initial state |
| -1 1 | α – – – C | first feature reached |
| 0 0 | α C – – C | pause; grid cells auto-reset |
| 1 1 | α C 1 1 – | move & track displacement |
|  | . . . |  |
| 3 3 | α C 4 4 B | feature B reached – store vector |
| 0 0 | α B – – B | pause; grid cells auto-reset |
|  | . . . |  |
| -2 -3 | α B -2 -3 – | mid-way to next feature |
|  | . . . |  |
| -3 8 | α B -5 5 A | feature A reached – store vector |
|  | . . . |  |
| 13 -6 | α A 8 -1 D | feature D reached – store vector |
|  | . . . |  |
| -8 8 | α D 0 7 E | feature E reached – store vector |
|  | . . . |  |
| -7 -8 | α E -7 -1 C | feature C reached – store vector |

*PL Contents*

| | |
|---|---|
| α B -5 5 | **A** |
| α A 8 -1 | **D** |
| | |
| α E -7 -1 | **C** |
| α C 4 4 | **B** |
| α D 0 7 | **E** |

**Figure 5. Exploration of environment α. The blue arrows indicate the state update sequence: given a state vector, a movement is applied, and this leads to the next state.**

| e t Δx Δy | head |
|---|---|
| α B -5 5 | A |
| β E 3 4 | A |
| β B 5 0 | A |
| β A -9 -4 | C |
| α E -7 -1 | C |
| α C 4 4 | B |
| β C 4 4 | B |
| α A 8 -1 | D |
| β B -6 1 | D |
| β D 8 -5 | E |
| α D 0 7 | E |

**Figure 6. Contents of PL after both environments α and β have been explored. Contents are organized according to *head*; for a given *head*, entries are associative. Abbreviations: *e*: *eId*, *t*: *tail*.**

*Orientation and Navigation*

Assume orientation begins with *PL* contents as in Figure 6. The orientation sequence is in Figure 7. The mouse is placed in the β environment at location [5,5]. At that point, it knows nothing regarding its whereabouts, so the *eId* is the union of all the environments (in this case there are only two). The state vector is otherwise *null*. The agent then begins moving about the environment like it did during exploration. To shorten the example, say that it happens to make the move [-1,1] and encounters feature C, and the *state vector* becomes [α,β - - - C]. Because the proximal input is non-*null*, this is sufficient to produce *i_eId* ← *PL*([*state vector*]). All PL entries with C as the *head* match *state vector* equally well. After a single cycle pause for the *tail* to be updated with the *head*, both *i_eId*s remain part of an ambiguous state vector [α,β C - - C].

The orientation process continues through a series of movements summarized by the movements [-2,-2], and [6,6] so the net movement is [4,4], and feature B is encountered. When *PL* is queried with [α,β C 4 4 B], it happens that both environments have exactly the same edge, so *PL* again returns an ambiguous *i_eId*. The orientation process continues, and after a net movement of [-6,1] a third feature D is reached, and the state vector becomes [β B -6 1 D]. This time when *PL* is accessed, the *i_eId* is unambiguous because only environment β has a completely matching edge. At this point orientation is achieved. The agent knows it is in environment β at feature D.

After a one step pause, navigation begins. The state vector is [β D - - D], and the agent performs a *PL* query to infer displacements that will take it to feature E. The agent sets *target* ← E, and *i_Δx*, *i_Δy* ← *PL*([β D - - E]). The best match yields *i_Δx* = 8, and *i_Δy* = -5. As can be seen in Figure 4, these displacements will, in fact, take the agent to feature E.

| movement | state vector | comments |
|---|---|---|
| - - | α,β - - - - | initial state – any environment is possible; otherwise *null* |
| -1 1 | α,β - - - **C** | first feature reached; environment is ambiguous |
| 0 0 | α,β C - - C | pause, access *PL* |
| | | |
| -2 -2 | α,β C -2 -2 - | perform random movement to a featureless grid cell |
| | . . . | |
| 6 6 | α,β C 4 4 B | continue until second feature reached; environment is ambiguous |
| 0 0 | α,β B - - B | pause, access *PL* |
| | . . . | |
| -6 1 | **β** B -6 1 D | third feature reached; environment is β;  orientation complete |
| 0 0 | β D - - D | pause – then navigate to feature E |
| 0 0 | β D - - **E** | Apply feature E to *PL* to get inferred Δ*x*, Δ*y* |
| **8 -5** | β D 8 -5 E | inferred displacement becomes next movement |

**Figure 7. Example of orientation and navigation. Bold underlined values are inferred.**

## 3. Implementation: Coding and Synchronization

Given the state machine description as a starting point, the next few sections lay out a biologically plausible spike-based architecture that implements the state machine and associative memory.

### 3.1 Coding and Communication

At system interfaces, information is communicated via spikes modeled as bit vectors. Intermediate values internal to the model are often encoded as vectors of low precision non-negative integers.

Communication takes place over *bundles* of lines – essentially parallel buses. Vectors may be formed as the concatenation of one or more bundles.

A state vector is represented as a vector containing 5 bundles. As an example, say there are 4 environments, 4 features, and an 8×8 grid. If the state vector components are *eId* = 3; *tail* = 2; $\Delta x$ = 4; $\Delta x$ = 3; *head* = 4, a 1-hot encoding is:

|0010|0100|00010000|00100000|0001|.

### 3.2 Synchronization

High level neural operation is consistent with Abeles's *synchronous synfire chains* [1]. Neurons are organized in layers ("pools" according to Abeles) and processing proceeds as waves of spikes (volleys) pass from one layer to the next in a synchronized fashion. In a given volley, there is at most one spike per line.

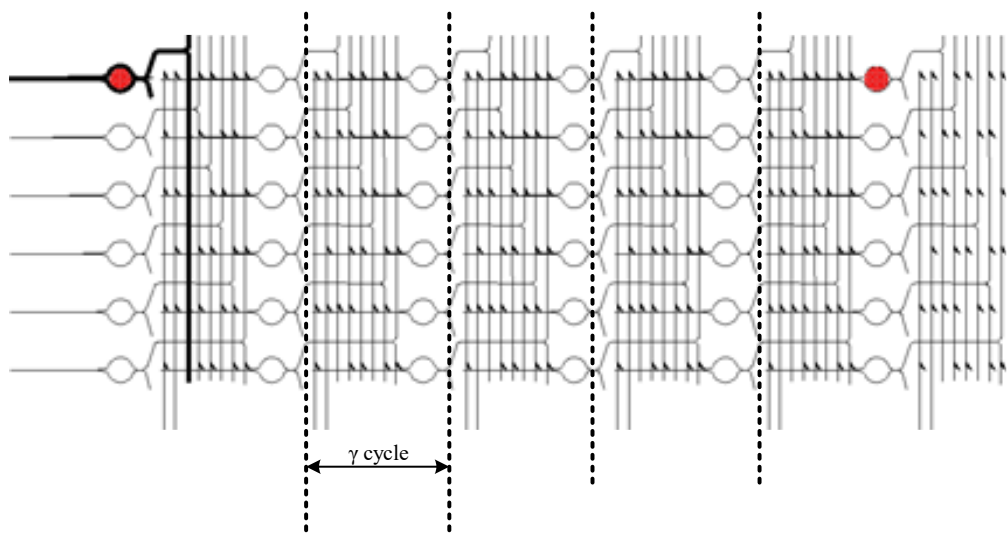

**Figure 8. An Abeles synfire chain organizes neurons (circles) into layers, separated by synaptic crossbars (small triangles). Spike volleys pass from one layer to the next, synchronized via gamma cycle inhibition. Each volley contains at most one spike per line. As pointed out by Abeles, the layout in the figure does not represent the anatomy, rather it represents the processing time sequence. Hence, the figure is drawn in an unrolled form for illustrative purposes; in reality, there can be any amount of feedback among layers. For example, the red neuron shown at two different points in the chain is the same physical neuron. (Figure taken from Scholarpedia [1])**

A *gamma* clock [4] marks off cycles that correspond to discrete computation *steps*. A gamma cycle is of sufficient length to cover the total time it takes to transmit a spike vector to the synapses of an excitatory neuron plus the time it takes for the synapses, dendrites, and neuron body to generate an output spike (if any) followed by a pass through winner-take-all (WTA) inhibition (not shown explicitly in the figure).

WTA inhibition coincides with the inhibitory portion of the gamma cycle [3]. In the remainder of the paper "step" refers to a gamma cycle.

## 4. Neurons with Active Dendrites

Before proceeding to Section 5 where macrocolumn architecture is described from the top down, this section first describes the basic element of computation, the active dendrite. Active dendrites, as modeled here, implement online clustering, and it is a fundamental tenet of this research that online clustering underlies all neural computation.

Spiking excitatory neurons composed of active dendrites are patterned after the HTM neuron model described in Hawkins and Subutai [9]. The spiking implementation defined here is not exactly equivalent to the HTM implementation but provides the same features and capabilities.

Briefly, the neuron body is fed by a number of dendrites, on the order of five to seven in a biological excitatory neuron, but more or fewer in the model neurons used here. Each dendrite is composed of a much larger number of segments or branches, each of which receives a number of distal inputs (see Figure 9). Also associated with each of the model dendrites is a proximal input. Although one could potentially model more than one proximal input per dendrite, in the model developed thus far there is only one. Synapses join the inputs to the dendrite, and each synapse has an associated weight. After learning, synaptic weights tend to be bimodal: weights are either close to 0 or a maximum weight, $w_{max}$.

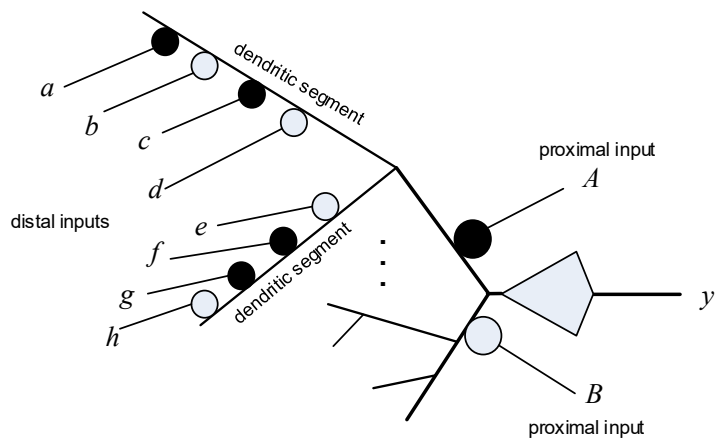

**Figure 9. Spiking excitatory neuron with active dendrites. The dark synapses have maximum weight, and the lightly shaded synapses have zero weight. Input spikes on *a*, *c*, and *A* will yield a relatively high total weight (reflected in the neuron's body potential) indicating a good pattern match. An input spike on *A* only (or input spikes on *b*, *d*, and *A*) will yield a lower body potential indicating a weaker pattern match; Input spikes on *a* and *c* without a spike on proximal input *A* yield no output spike.**

The biological rationale articulated in [9] is that a single spike arriving at a distal synapse is unable, by itself, to generate a pulse strong enough propagate along the dendrite and increase a neuron's body potential by any significant amount. Rather, it takes multiple distal spikes in close spatial and temporal proximity to generate a strong dendritic pulse that will bring the neuron's body potential to a level where the neuron is primed to spike. Even then a spike on a proximal synapse is required to trigger an output spike on the neuron's axon.

In the model, a spike on a proximal input is a necessary and sufficient condition for generating a neuron output spike. In a classic spiking neuron model the number of spikes arriving at maximum weight synapses determine how quickly the neuron's body potential rises. Consequently, the neuron's output spike occurs earlier or later, depending on how many distal input spikes arrive at maximum weight synapses. The more input spikes with maximum weight synapses, the higher the body potential and the earlier the output spike. In contrast, in this work the model neuron's output is the body potential, itself, rather than a temporal spike.

As a general interpretation, the pattern of distal input spikes defines a specific *context*, and a proximal spike indicates a specific *feature* that is present at the specified context. In the model defined here, distal inputs are divided into bundles with each bundle being associated with some part of the context.

For the mouse-in-the-dark benchmark, there are four distal input bundles: *eId, tail,* $\Delta x$*.* and $\Delta y$*.* The proximal input is *head,* the feature where the mouse currently resides. Following subsections describe excitatory neuron operation in more detail.

### *4.1 Dendritic Segments (Point Integrators)*

*PL* is an associative memory that stores state vectors in dendritic segments. Each segment stores a specific context encoded in the synaptic weights. An individual dendritic segment implements a *point integrator* as illustrated in Figure 10a. The point integrator is shown in "einschematic" form in Figure 10b. Einschematics are so-named because they draw their inspiration from "einsum" notation commonly used in machine learning frameworks. It is a simple, yet unambiguous, function specification in schematic form. Einschematic notation is summarized in Appendix 1.

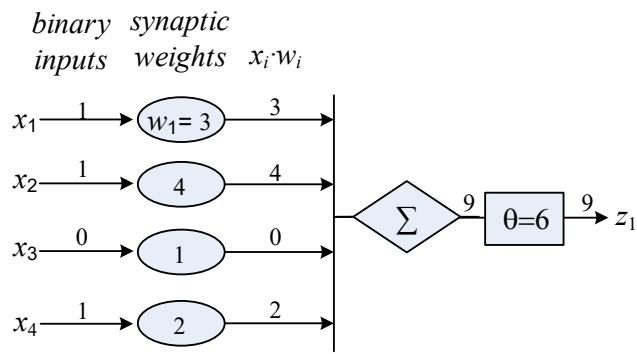

a) **Example point integrator**

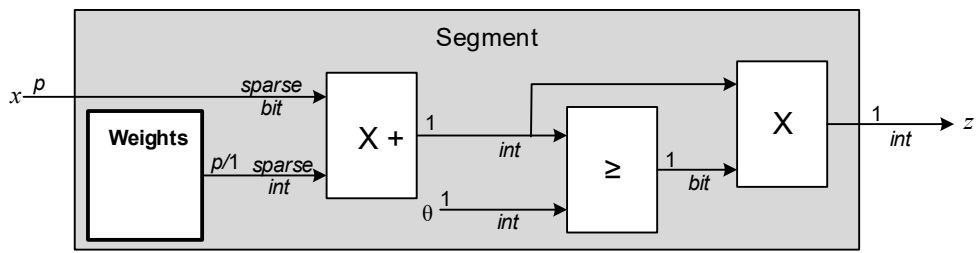

b) **Einschematic for point integrator**

**Figure 10. A dendritic segment is implemented as a point integrator. a) Distal inputs $x_1$ - $x_4$ are communicated over a bundle of 4 lines. As spikes, encoded as a bit vector, arrive at the synapses, a dot product with integer weights yields the body potential for the given segment. If the potential reaches the threshold value $\theta = 6$, the value of the potential is produced on output $z$. b) A general point integrator in einschematic form. Observe that when a multiplication takes place, one of the operands is always a bit or a bit vector, thereby simplifying the "multiplication" significantly. The annotations "int", "bit", and "sparse" are not required, but may be added to specialize the functional block implementation according to its inputs' properties.**

In prior work by the author, the temporal neuron model is a point neuron. In this work, a multi-point active neuron model is used so individual segments become point integrators. Also, in conventional temporal neuron models, the body potential is converted to a spike time. In the point integrator model used in this paper, this conversion does not take place, rather the potential, itself, is the output. This leads to an easier-to-understand model that, when combined with a WTA network (next subsection), is functionally equivalent to the temporal model. For this reason information internal to the model (i.e. body potential) is in the form of integer vectors, rather than being exclusively bit vectors. Thus, the model's function remains consistent with the temporal model, although the function specification is not explicitly temporal. (See Appendix 2 for a more detailed discussion of this and related issues)

### 4.2 WTA Inhibition

WTA inhibition is a key component of a complete dendrite (see Figure 11). Inhibitory neurons are not modeled individually. Rather they are modeled as a bulk WTA process. WTA methods go far back in the annals of machine learning, and Thorpe [30] first proposed that WTA inhibition should be performed on a bundle of lines carrying temporally encoded spikes.

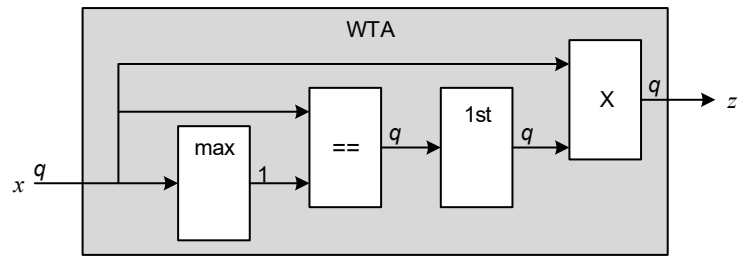

**Figure 11. Schematic for 1-WTA inhibition. The maximum value of a size $q$ integer vector $x$ is found. The output vector $z$ contains 0s in every component except the component with the maximum $x$ value, in that component $z$ is assigned the value in $x$. With 1-WTA, if there is a tie, then the component with the lowest index is selected (via the "1st" block). This is a systematic method that makes simulator debugging easier. However, in general the "winner" of a tie can be arbitrarily chosen, including via pseudo-random selection.**

A WTA inhibition block's input and output bundles have the same number of lines. An integer input vector on the input lines is translated to an output vector containing only the element with the highest value. Tie cases are handled in one of two ways (although an eventual research objective is to reduce this to one). Most commonly they are handled as 1-WTA described in the Figure 11 caption. However, an alternative to be used in Section 0 allows all tying elements to pass through WTA inhibition (t-WTA).

### 4.3 Dendrites

A complete dendrite (Figure 12) consists of multiple segments, each implemented as a point integrator, followed by WTA inhibition. The final output is enabled by a proximal input. The proximal input becomes a *de facto* enable signal because it is a necessary and sufficient condition for the integrator to emit an output spike.

In the research benchmark, each dendrite is associated with a specific state vector *head*, and the segments track the contexts in which the given *head* feature appears.

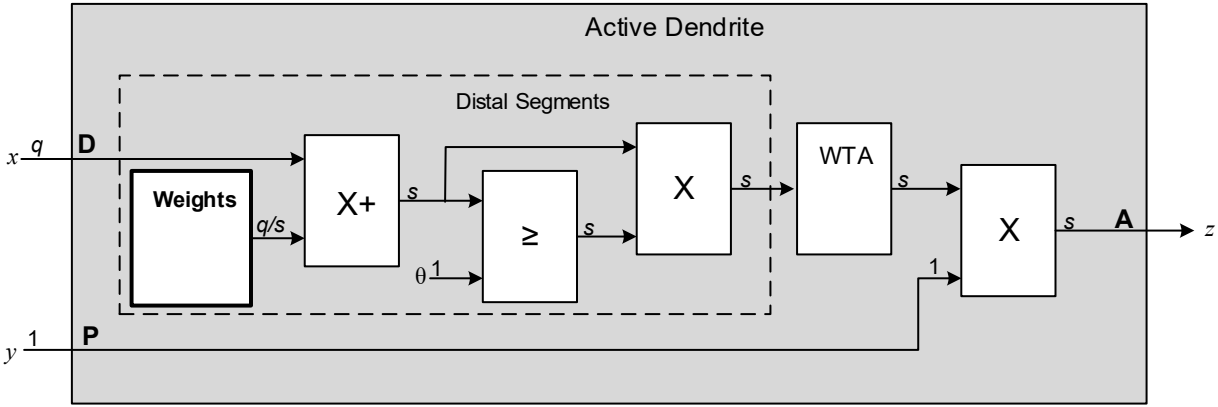

**Figure 12. An active dendrite containing *s* segments (point integrators enclosed in dashed lines) that operate on distal inputs (D). The point integrator outputs pass through WTA inhibition followed by multiplication by a 1-bit proximal input (P). The dendrite's final output (A) is a scalar which is the highest potential computed by any of the segments.**

### *4.4 Excitatory Neuron*

The schematic for the inference portion of an excitatory neuron is in Figure 13, although the distal input structure to be used later is a little more complicated than this. The neuron consists of any number of active dendrites (*q* in this case) whose outputs are all *max*-ed together. Each of the dendrites is associated with a different proximal input, which specifies a feature, and they all share the distal inputs, which specify contexts.

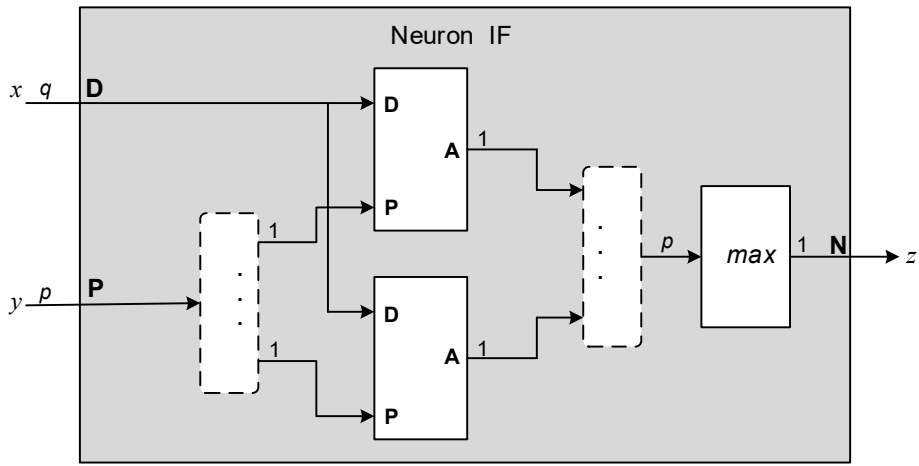

**Figure 13. An excitatory neuron consisting of *p* dendrites that share a distal input bit vector (D). The proximal input (P) is a bit vector containing *p* elements, one for each of the dendrites. The outputs of the individual dendrites are integer vectors, the maximum of which is passed through to the neuron's output (N), a scalar integer.**

### *4.5 Spike Dependent Plasticity (SDP)*

State vectors are stored in the associative memory *PL* via an online learning process. As noted above, state vectors are organized into segments according to the proximal input *head.* Storing state vectors in *PL* segments is not a one-shot process as in the abstract associative memory described above. Rather, they are stored via a learning process where multiple applications may be required before the state vector is solidly entrenched in a segment.

Because a proximal input spike is necessary and sufficient for dendritic activity, all the proximal synapses that receive input spikes quickly converge to $w_{max}$. Hence, to simplify modeling, they are implemented as a binary enable signal.

When temporal neurons are modeled, spike timing dependence plasticity (STDP) is commonly used for online learning [2] [7]. In this document the temporal aspect has been abstracted away so inputs and outputs are binary vectors (one bit of temporal precision). Note that the following discussion applies to point integrators – either dendritic segments or point neurons. The SDP update functions for distal synapses are given in Table 1. Weights saturate at 0 and at $w_{max}$. Although SDP learning is not the focus of this document, the motivation behind the method merits some explanation. First, consider the way that *backoff* and *capture* control the formation of clusters. In particular, the relative values of *capture* and *backoff* affect the distance (or alternatively *overlap*) between members of the same cluster. That is, once a cluster is established and a new input bit vector is applied, whether the new vector should become part of an existing cluster or should begin a new cluster is related to the overlap.

**Table 1. SDP update function for distal synapses.**

| $x_i$ | $z_j$ | weight update | limitation |
|---|---|---|---|
| 0 | 0 | $\Delta w_{ij} = 0$ | |
| 0 | 1 | $\Delta w_{ij} = -backoff$ | $w_{ij} \geq 0$ |
| 1 | 0 | $\Delta w_{ij} = +search$ | $w_{ij} \leq w_b$ |
| 1 | 1 | $\Delta w_{ij} = +capture$ | $w_{ij} \leq w_{max}$ |

To illustrate the way capture and backoff interact, consider the situation where all the bit vectors contain exactly *m* spikes. (In practical cases this can be relaxed somewhat, but vectors should have approximately the same number of spikes). Say the *overlap* is the minimum number of bits that two vectors can have in common in order for them to be assigned to the same cluster.

Begin with an un-trained system where all the weights are at some baseline value $w_b$. When the first vector $x^1$ is applied, all the segments in a dendrite will have a potential of $pot(x^1) = mw_b$, and there is a WTA tie. Arbitrarily segment $z_i$ is selected and cluster $C^i$ is established. Because the input vector has *m* ones, *m* of $z_i$'s input synapses are increased by *capture*, and the remainder are decremented by *backoff*. As shorthand in the following equations, let variable *v* represent *overlap*, *c* represent *capture*, and *b* represent *backoff*.

Say the objective is for a subsequent vector to be placed in $C^i$ if it overlaps the first by at least *v* spikes; otherwise, it should be captured by a different segment $z_j$ to establish cluster $C^j$. At exactly overlap *v*, the total potential for segment $z_i$ is:

$pot(z_i) = (w_b + c)v + (w_b - b)(m-v)$
$= m w_b + cv - b(m-v)$

Meanwhile the potential of all other segments, including $z_j$, remain at their initial values:

$pot(z_j) = mw_b$

For cluster $C^i$ to capture the second vector if the overlap is *v* or less, it is required that:

$pot(z_i) > pot(z_j)$, or
$m w_b + cv - b(m-v) > m w_b$, or
$b(m-v) < cv$

If we fix $c = 1$, then

$b(m-v) < v$, or

$b < v/(m-v)$.

As a simple example, if the $m = 64$ and the desired overlap to be in the same cluster is $v = 48$ or more, then the *backoff* should be less than 48/16 or 3 when *capture* = 1.

The *search* parameter allows clusters to adapt to changing inputs. Once a segment $z_i$ has captured a cluster, but the input stream changes so that the cluster no longer reflects the current input stream, $z_i$ may no longer spike. Whenever a synapse receives an input spike, but there is no output spike, the synaptic weight will be incremented by *search* (up to some maximum weight $w_b$, where typically $w_b < w_{max}$). In practice *search* << *backoff* and *capture*. So all the synaptic weights for $z_i$ will gradually creep upward in accordance with the new input patterns as long as $z_i$ does not spike. Eventually, the weights will be high enough that $z_i$ spikes and a new cluster is captured. Or, if a member of the original cluster appears, $z_i$ will spike and *backoff* will decrement the weights of the searching synapses.

### 4.6 SDP

A schematic for the *capture* update is in Figure 14. The weight matrix dimensions are $p/q$. First, there is a matrix multiplication of input $x$ transposed, a $p$ bit column vector, and output $y$, a $q$ bit output row vector, yielding a $p/q$ bit mask. The bit mask is multiplied by the *capture* value, and this is added to the weights. Limiting the weight to $w_{max}$ is done by the *min* function at the end.

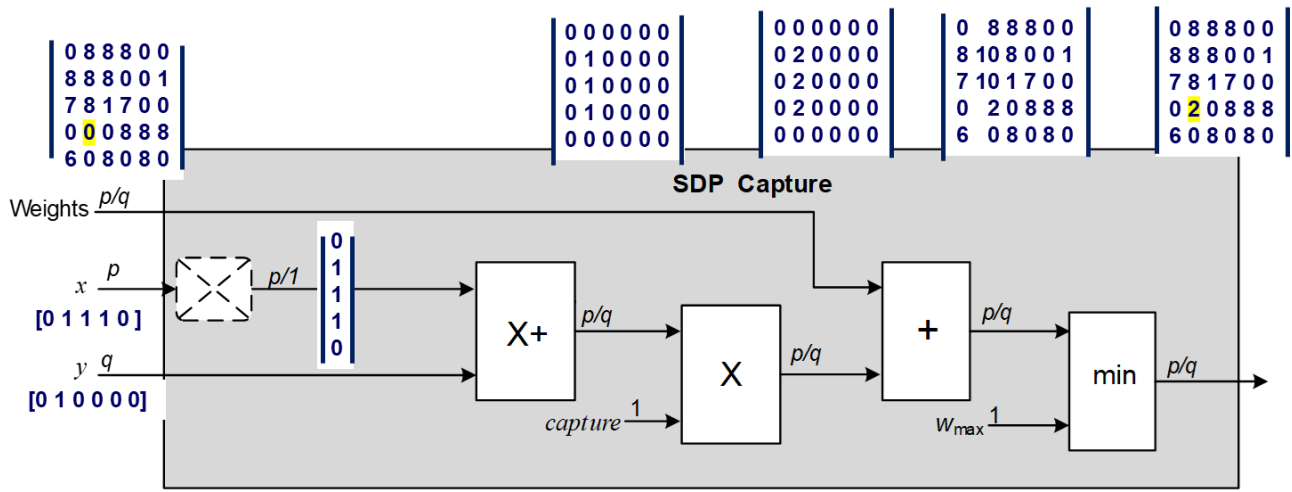

Figure 14. Schematic for SDP capture update. Example vectors and arrays for a typical update are shown.

All three of the update functions – *capture*, *backoff*, and *search* – can be done in a similar way with *backoff* performing a *not* function on the $x$ input, and *search* performing a *not* on the $y$ input. These three update functions are mutually exclusive so they can be done in parallel with their outputs merged before being added to the *weight* array. This and other details of the design are discussed in Section 7.3.

### 5. Macrocolumn Architecture from the Top Down

Figure 15 is a system schematic that consists of a macrocolumn, an agent, and a model environment. The state vector (SV) is shaded in white. The macrocolumn's function is divided into two primary sub-functions.

*Grid cells* (**GD**) – implements a 2-d grid and path integration.

*Place cells* (**PL**) – implements an associative memory as described above.

*GD* and *PL* are connected together and with the *agent* via *select blocks* (cS and dS) described below. Synchronization follows the Abeles synfire chain model (Section 3.2). The gamma cycle boundary is shown as a vertical dashed line. At the gamma cycle boundary, state vector components are synchronized with data buffers (B) that function as clocked D latches.

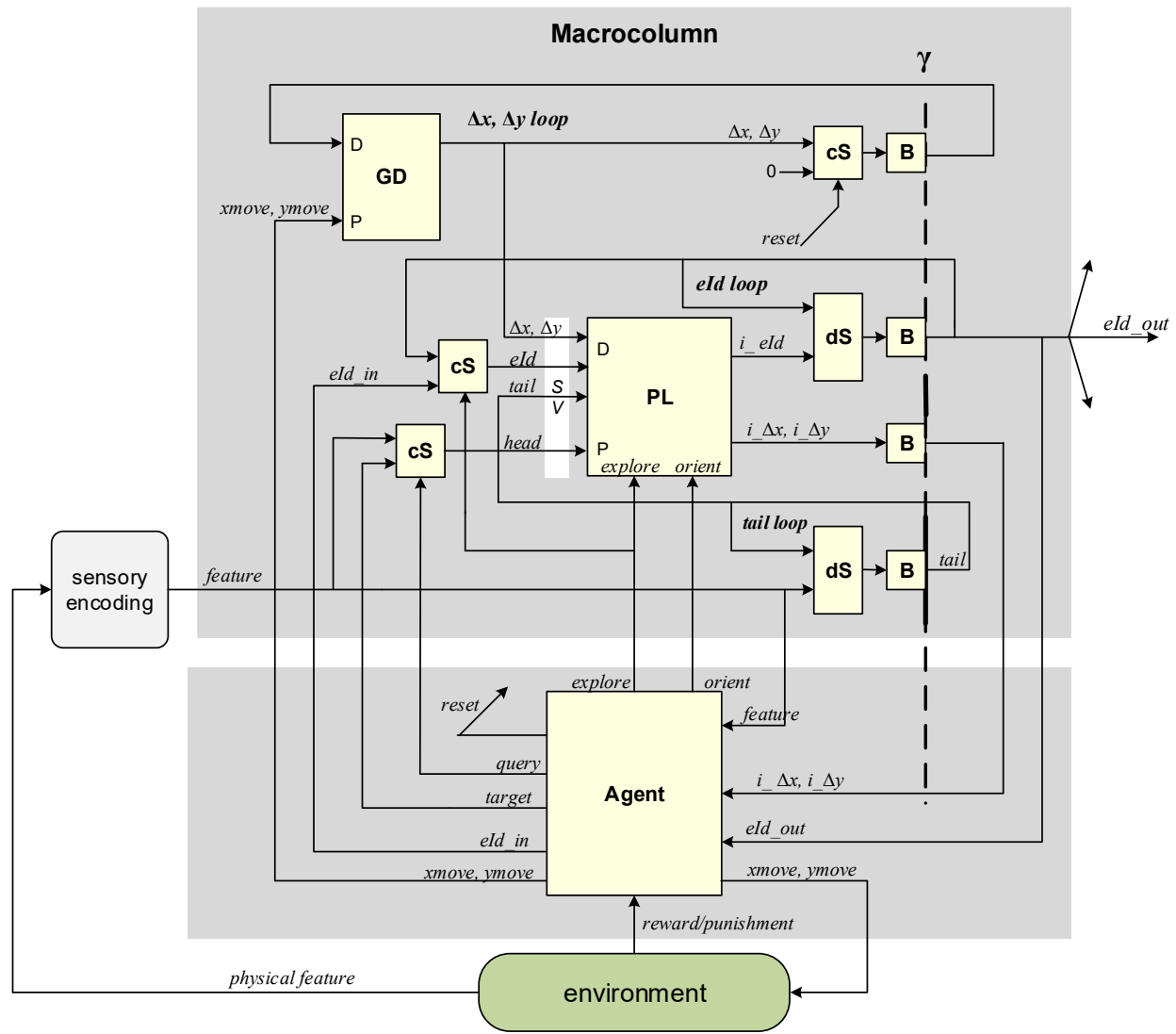

**Figure 15. System schematic. Agent and environment blocks are part of the simulation framework. The state vector (SV) is shown at the inputs of PL.**

## 5.1 Connecting Blocks

The select blocks (cS and dS) are essentially multiplexors that select from one of two inputs (Figure 16). For state vector update operations that depend on a mode, these select the mode-dependent inputs. The *control select* (cS) (Figure 16a) selects one of two inputs under control of the agent via a two-rail select signal. A [1 0] select input sends the input D0 to the output; a [0 1] selects the D1 input. The *data select* (dS) (Figure 16b) selects the D1 input if it is nonzero; otherwise the D0 input is selected.

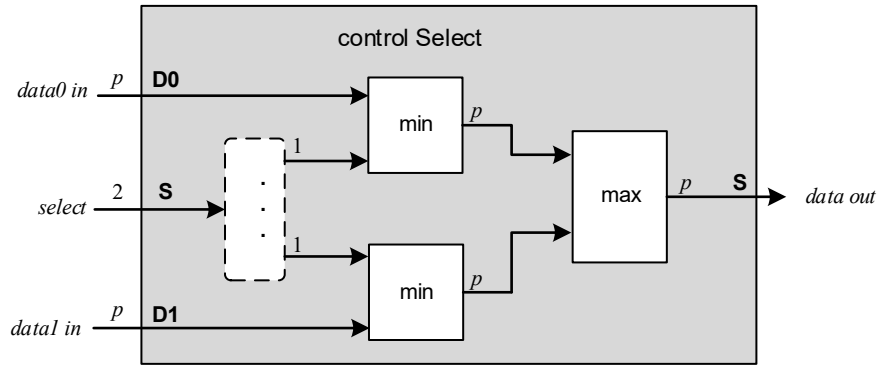

a) cS block.

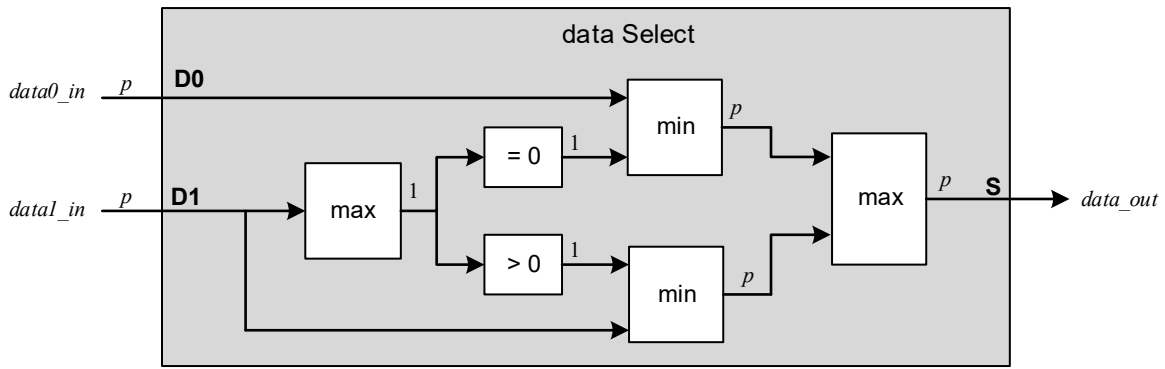

b) dS block.

Figure 16. Select functions.  a) The cS block acts more-or-less as a conventional multiplexor with a 1-hot control input. b) The dS block selects the D0 input if the D1 input is null; otherwise the D1 input is selected; new data replaces old data.

### 5.2  State: Autaptic Loops

A guiding principle of biologically plausible spiking neural computation is that only synaptic weights hold static state, i.e., state that persists in the same place for relatively long periods of time (many gamma cycles).  There is another way to maintain long term state that is not static. It is well documented that a significant number of neuron connections are *autaptic* [22] – that is, the axon of a neuron feeds back to one of its own synapses.  This forms an autaptic loop that provides a dynamic mechanism for holding state variables.  The macrocolumn in Figure 15 contains three autaptic loops. These hold the *tail*, *eId* , and displacement ($\Delta x$ and $\Delta y$) components of the state vector.

*Tail Loop*

Figure 17 is the autaptic loop for the *tail* state variable.  It implements the state update function:

      if *head* ~= *null*  *tail* ← *head*
      else *tail* ← *tail*

As long as head remains *null* the same *tail* value will circulate around the loop indefinitely. Eventually, if a non-*null head* value comes along, it replaces the current *tail*, and the new *tail* is captured in the loop.

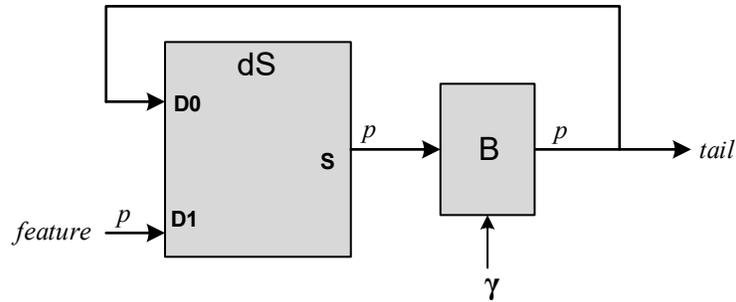

**Figure 17. Autaptic loop that holds the *tail* state variable.**

*eId Loop*

The *eId* loop implementation is similar to the *tail* loop implementation. The only wrinkle is that the *eId* loop performs environment disambiguation during orientation, so the loop may contain unions of environments rather than a single environment. This means the *eId* encoding is not necessarily 1-hot. It is *k*-hot if the actual environment has been narrowed down to *k* possibilities.

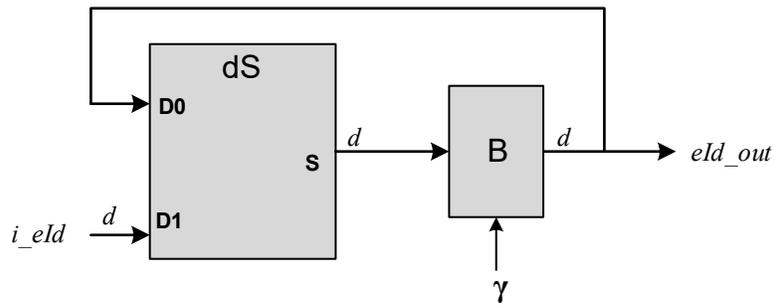

**Figure 18. Autaptic loop that holds the *eId* state variable.**

Reducing multiple *eIds* to a single *eId* is a straightforward form of disambiguation as put forward in [14]. This is illustrated in the example of Figure 7. In that example, *PL* is initially queried to determine *i_eId*. If there are multiple entries with the same number of state variable matches, then WTA produces multiple non-zero outputs. These enter the *eId* loop by the same process used in the *tail* loop. As the agent moves on to new features, all the components of the query are filled so a complete state vector is available for matching during a query. This means that only environments with exactly the same edges in their graphs pass through a narrowing process. Unless the graphs are identical, this process will eventually result in a single unambiguous *eId*.

*Displacement Loop*

The displacement loop tracks the *x* and *y* grid displacements $\Delta x$ and $\Delta y$ from the most recently visited feature. It is reset under control of the agent, but otherwise it simply holds the most recently computed displacements through the gamma cycle.

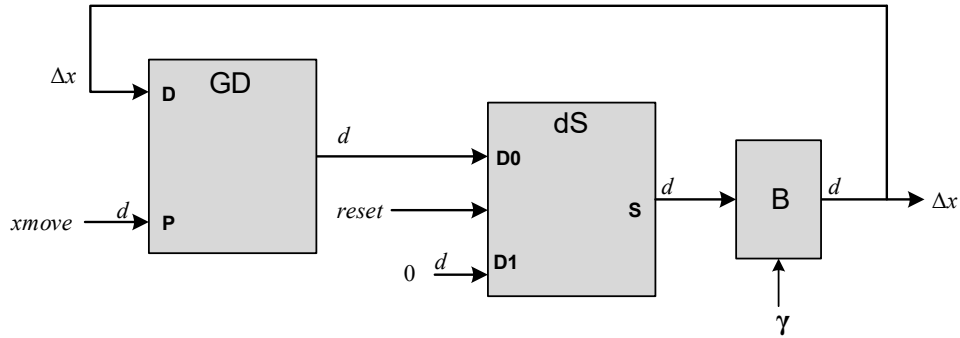

**Figure 19. Autaptic loop that holds the Δ*x* state variable. The Δ*y* loop is similar. The reset signal is applied whenever a feature is reached.**

## 6. Grid Cells and Path Integration

In the system studied here, the agent invokes actions that cause physical movement through a 2-d environment. Simultaneously, via *path integration*, the Grid cells (*GD*) track an analogous movement through an internal grid representation.

Because "all thinking is movement", when higher level thinking is taking place, the only movement may be internal. Furthermore, even when navigating a physical environment, internal movement can be invoked without the corresponding external actions. For example, virtual movement or "mind travel" [23] can be used by an agent to plan future actions to achieve some objective.

The application of grid cells and path integration to high level thinking is clearly a very challenging problem. By choosing the problem of physical navigation through 2-d space, many of the big challenges are admittedly being sidestepped in favor of demonstrating the basic elements of a complete working system.

As illustrated in Figure 15, grid cells and path integration are implemented with their own nearly autonomous state machines. Conceptually, grid cells and path integration act as a sort of trip odometer that is reset every time a new feature is sensed and, as movement occurs, it tracks net relative displacements (Δ*x* and Δ*y*) until the next feature is reached, at which time it resets and tracks movements to the next feature, etc. The displacement state is maintained in an autaptic loop.

### *6.1 Grid Cells*

For the research benchmark, there are two environments: an external physical one and an internal one in the macrocolumn grid used for tracking movement. The fact that both are 2-d spaces greatly simplifies the implementation.

Note that the biological grid system appears to be triangular [19], possibly because physical macrocolumns can be packed more tightly if a triangular grid is used. However, for the problem at hand there does not appear to be any computational advantage in using triangular grids, so a Cartesian system is chosen here to simplify the implementation.

In the *external space* considered in the research benchmark, locations and displacements are specified in a two-dimensional coordinate system (although the concepts extend to a space of any number of dimensions and/or coordinate system.) Agent-invoked movements through the environment occur in the external space. Specifically, *external movements* are reflected as changes in external space coordinates.

Displacements in the *internal space* maintained in *GD* are encoded as two spike volleys, each corresponding to one of the dimensions. Agent actions (the 2-d movements) are also specified with two vectors *xmove* and *ymove*. Both displacements and movements use one-hot encodings. For example, in an 8×8 system, the displacement [3,5] may be represented as two concatenated bit vectors: |00100000|00001000|.

## 6.2 Path Integration

An important observation [12] is that all biological macrocolumns implement an internal grid-based navigation system that appears to be effectively hard-wired. That is, it appears that an innate grid organization squeezes through the "genomic bottleneck" [35] and is a built-in mechanism throughout the neocortex. Some characteristics may differ from one region of the neocortex to the next, but the working hypothesis is that in every region there is an innate grid capable of holding frames of one type or another.

With one-hot encoded vectors, movements appear as shifts of the displacement vectors, so the model *GD* cells operate functionally as shifters. Although the *GD* cells can be implemented via neurons, in the model studied here the shift function is implemented directly in the simulator.

Recall that in the state vector implementation, $\Delta x$ and $\Delta y$ update functions are:
    if *feature* ~= *null*  $\Delta x \leftarrow$ *null*; $\Delta y \leftarrow$ *null*
    else $\Delta x \leftarrow \Delta x +$ *xmove*;  $\Delta y \leftarrow \Delta y +$ *ymove*

Referring to Figure 15, the *GD* block consists of two wrap-around shifters where the shift counts are applied to the P input and the displacements to be shifted are applied to D inputs. The shift output is held in an autaptic loop which feeds back to the inputs. Shifts wrap-around so the grid becomes a biologically plausible 2-d torus [5][23], and both positive and negative movements can be accommodated by the same shifter.

When there is a feature at the current grid location, the displacements are initialized to *null*. As movement occurs from a grid cell containing a feature to featureless grid cells, shifts of $\Delta x$ and $\Delta y$ track the agent-directed movements (*xmove* and *ymove*) by performing what are essentially unary wraparound additions.

The capabilities built into the specific shifter(s) depend on the modeled movements. For example, if movements are restricted to at most one grid cell per step, then only single step shifts are implemented. For the 2-d system considered here, shifts of any number of grid cells are allowed. This makes motion (and simulations) faster, although in an implementation it will be more expensive.

## 6.3 Efficient Grid Cell Encoding

Grids are based on a 2-d coordinate system. Each of the dimensions can be 1-hot coded, as just described, and this is fine for relatively small systems – 40 x 40, say. But if we want something orders of magnitude larger, say 500 x 500, then 1-hot coding becomes very inefficient and potentially expensive if a large set of shift amounts are implemented.

By example, a solution for the 500 x 500 case is to encode the value of each dimension with three 1-hot fields rather than one. To do so, select three field lengths that are relatively prime – 7,8,9, for example. There are 7x8x9 =504 unique codings using 7+8+9 = 24 lines. This is 3 out of 24 sparsity. If both dimensions are encoded this way a total of 48 lines are sufficient for accessing a 504 x 504 space.

With this scheme, movement of *d* steps along a dimension is performed by modulo shifting all three 1-hot codes by *d*. Because the field sizes are relatively prime, single step movements cycle through all 504 code combinations.

## 7. Place Cells (PL)

For the research benchmark, the complete *PL* (Figure 20) contains three minicolumns. One for each of the three state vector components that are accessed during orientation (*eId*) and navigation ($\Delta x$ and $\Delta y$). This design uses multiple-dendrite neurons having a dendrite for each of the proximal inputs.

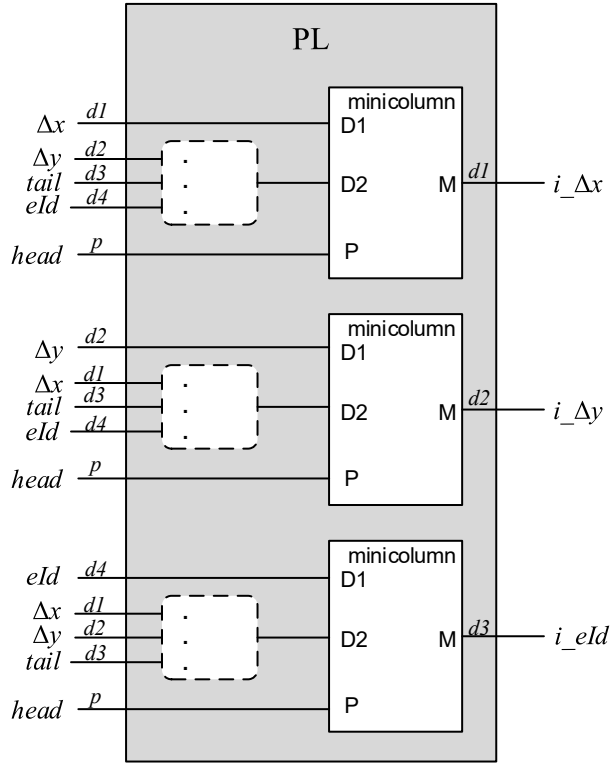

**Figure 20. Full *PL* implementation for research benchmark application.  In this example, multi-dendrite neurons are organized as three minicolumns.**

## *7.1  Minicolumn*

The minicolumn used in this work (Figure 21) consists of multiple neurons feeding WTA inhibition.  The minicolumn implements a block of associative memory as described in Section 2.2. For each feature specified by a proximal input it associates a set of contexts where the given feature appears. The contexts are specified by distal inputs. In the research benchmark, each of three minicolumns holds a component of the state vector that is associatively accessed.  These are Δ*x,* Δ*y,* and *eId.*   The complete set of state variables also includes the *tail* and the *head*.  These two are part of the context, but they are never retrieved as part of the inference process.  Hence there are three minicolumns in *PL*.  For all three, the proximal input in the research benchmark is *head*.

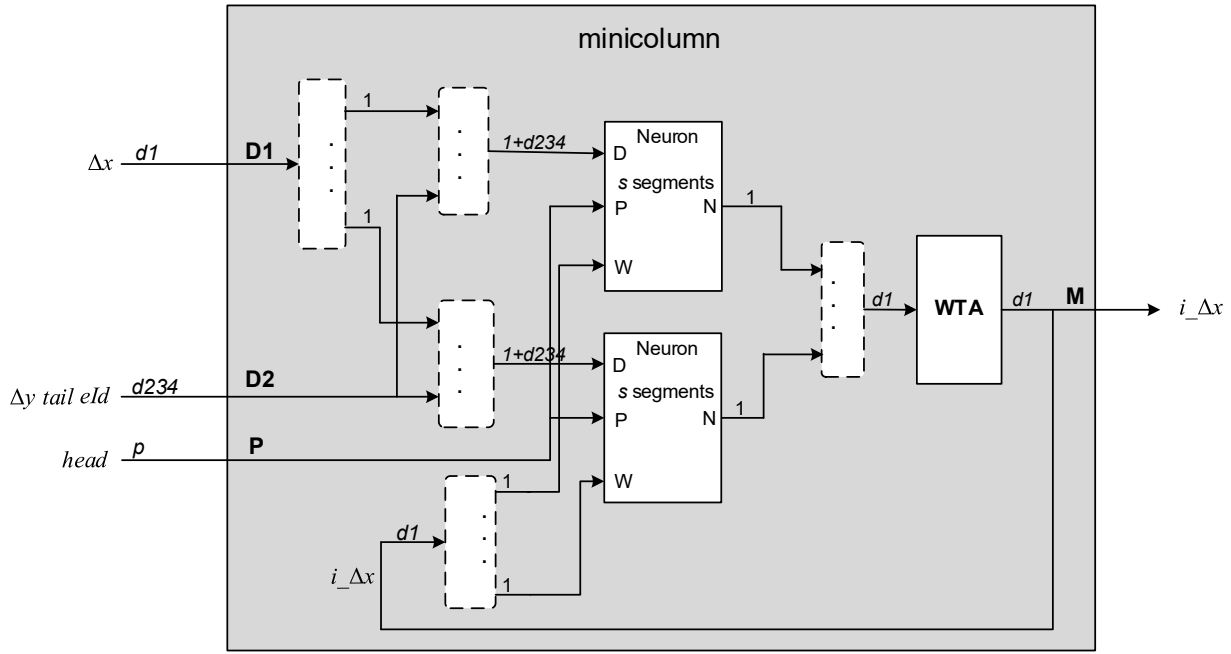

**Figure 21.** *PL* is composed of multiple minicolumns. For the working example, the minicolumn for state variable Δ*x* is illustrated here. There is a neuron for each bit of the Δ*x* state vector. The feedback of *i_Δx* passes through synaptic weights which are updated at the end of each gamma cycle before contributing to the subsequent *i_Δx* (see Section 7.2).

In the figure, the minicolumn is for the *inferred* variable Δ*x*, so the output is labeled "*i_Δx*". One of two distal inputs is Δ*x*. Other state variables Δ*y, tail,* and *head* in combination form a second distal input. The reason for partitioning the distal inputs this way is that there is no functional reason the distal input Δ*x* and the output *i_Δx* must use the same encodings; layers of processing separate them. In other words, when the learning process proceeds in a natural way the encodings are determined more-or-less independently. For implementation simplicity and ease of understanding, however, we would like the input Δ*x* and the *i_Δx* output to use the same encoding. This can be guaranteed by forcing the *i*th bit Δ$x_i$ to be the only distal input from Δ*x* that can be stored in a neuron segment that produces the output *i_*Δ$x_i$. Hence, there are two dendritic bundles D1 and D2 that are merged inside the minicolumn. One is special for Δ$x_i$, for the reason just noted. The other contains the concatenation of other distal inputs.

## 7.2 Excitatory Neuron

A minicolumn is composed of one neuron for each bit of the state vector that it is responsible for. A neuron for bit Δ$x_i$ is in Figure 22 and is divided into two components one of is responsible for inference and the other is responsible for updating synaptic weights as part of the learning process.

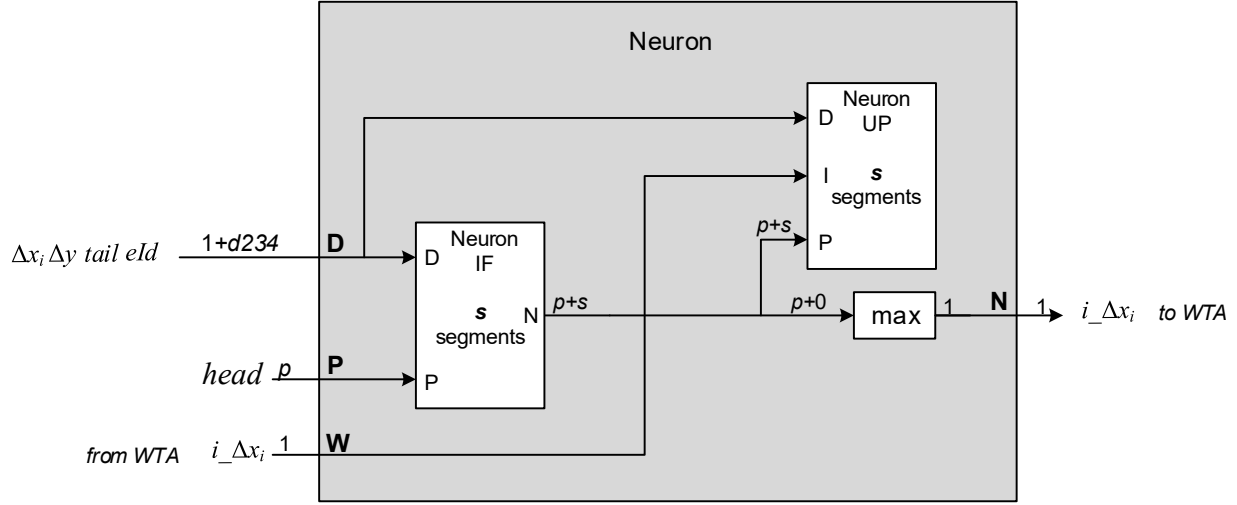

**Figure 22. A model neuron is composed of two components: one performs inference, and the other performs synaptic updates. They are conceptually separated because they are physically separated by minicolumn level WTA which is external to the neuron.**

### 7.2.1 Neuron Inference

The neuron subblock responsible for inference is shown in Figure 23. Every cycle a minicolumn receives an input *state vector* and performs inference. Some of the *state vector* components may be *null*, but the *head*, specified by a proximal input, must be non-*null* for inference to take place. The inference process produces a stored *state vector* that best matches the non-null components of the applied *state vector*. For the three minicolumns:

$$i\_\Delta x, i\_\Delta y \leftarrow PL(state\ vector)$$
$$i\_eId \leftarrow PL(state\ vector)$$

These inferences are used in two different ways.

The displacements $i\_\Delta x$ and $i\_\Delta y$ are inferred when making navigational queries. Informally, the query is: "I am at feature A and want to get to feature B. What $\Delta x$ and $\Delta y$ will get me there?" For this type of query, only the *eId* and *tail* components of the *state vector* are non-*null*, and the agent sets the *head* to a *target* feature. Hence, only the *eId*, the current feature (the *tail*), and the desired *target* are given. After passing through WTA inhibition, the $\Delta x$ and $\Delta y$ of the best matching state vectors are returned as 1-hot inferred outputs $i\_\Delta x$ and $i\_\Delta y$. Furthermore, the output also includes the segment that yielded the best match. This will be needed for synaptic updates because there are two levels of WTA, one in the neuron as shown here, and another in the minicolumn. In order to perform synapse updates using SDP, the segment that initiated an output spike must be known. The segment identifier is selected by a *p* channel cS block (a *p* input multiplexor with a 1-hot control input).

The *environment Id* is inferred during orientation. As the agent roams an unknown environment the *eId* encodes the union of potential environments. Hence, *eId_out* is not necessarily 1-hot; it can be many-hot, with each of the spikes identifying one of the possible environments. So, when orientation begins the agent sets all the bits in *eId_in* because any of the environments is possible. Consequently, as the agent roams during orientation, the *eId* is always non-*null*, as are the three other distal state vector components: the *tail* contains the most recent feature, and $\Delta x$ and $\Delta y$ contain the displacement from that feature. When the agent is between features, the *head* is *null*. As soon as a feature is reached, the *head* becomes non-*null* and the resulting proximal spike enables inference. The inference process yields an $i\_eId$ that is the union of the input *eIds* that are most consistent with the other distal components of the state vector. This is a subset of the input *eIds*, and in general it will be a proper subset that weeds out some of the potential

*eIds* contained in the input union. Ideally, as the orientation process proceeds the sequence of proper subsets eventually yields a unique *i_eId*. At that point, orientation is complete: the agent is at a known feature located in a known environment. In a less ideal situation, if the exploration process does not learn the environment thoroughly enough or two or more environments are identical, then the *eId* may remain ambiguous.

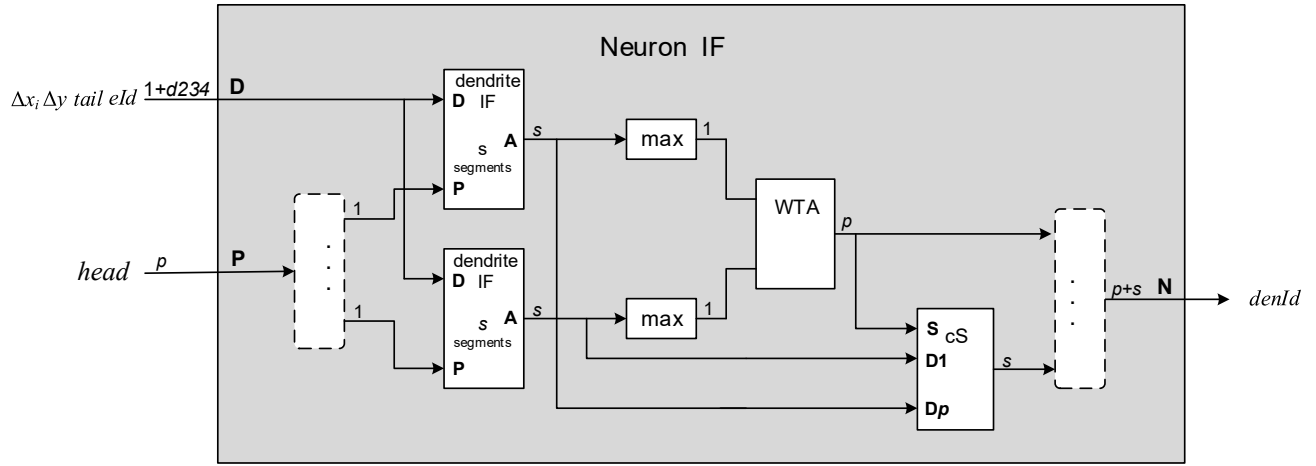

**Figure 23. A neuron inference block contains a dendrite for each proximal input (*head*). Each dendrite contains *s* segments.**

*Example*

In the research benchmark, inputs applied to excitatory neurons are [*eId tail Δx Δy head*], where *head* is a proximal input, and the others are distal inputs. Consider inference occurring during navigation (see the last entry in Figure 7). The state vector contains a known *eId* β and a known *tail* D (the feature where the agent currently resides); the displacements are *null*. The agent would like to find the displacements to feature E. Therefore, it performs a query by setting the *head* to the *target* (E) and applying the *state vector*: [β D - - E] to the *PL* inputs (and therefore the minicolumns that hold the displacements). Assume learning has been completed and all weights are at either 0 or $w_{max}$ = 8. Referring to Figure 6, there are two state vectors that have E as the head ([α D 0 7 E] and [β D 8 -5 E ]). The second state vector [β D 8 -5 E] has two distal matches β and D. Assuming their weights are $w_{max}$ = 8, the dot product is 16. The other state vector has one distal match and yields a dot product of 8. Therefore, due to WTA, *i_Δx*, comes from the best matching state vector [β D 8 -5 E], so *i_Δx* = 8. Similarly, the *i_Δy* minicolumn outputs *i_Δy* = -5. □

### *7.2.2 Synaptic Update*

The neuron subblock responsible for learning is shown in Figure 24. Control is very simple – there is an update at the end of every cycle – potentially a read/modify/write. In this model, all the weights for given dendrite are held in the same memory block. This is one of many ways of distributing the physical synaptic memory.

The most interesting part of the process occurs in the Dendrite Update sub-block, which is described below in Section 7.3.2.

In this block the weights are read under control of read/write (R/W) signals and two enable signals (E1 and E2). The enable E1 is activate if the subject neuron produced a spike output after minicolumn level WTA. The enable E2 depends on whether this was the dendrite that initiated the spike that passed through dendrite level WTA. If so, weights in the given dendrite should be updated. For a dendrite containing *d* distal inputs and *s* segments, there are *d\*s* synapses conceptually arranged in a *d×s* array.

Potentially, any of the weights could be updated, depending on what the distal inputs are. In this case, all the synaptic weights are read and sent to the dendrite update sub-block. However, if *search* = 0, which may be the case in some applications, then no updates need to be done for any of the segments except the one that caused the dendrite to spike. Consequently, the 1-hot segment identifer can be used to access a single column of weights, because these are the only weights that can be updated.

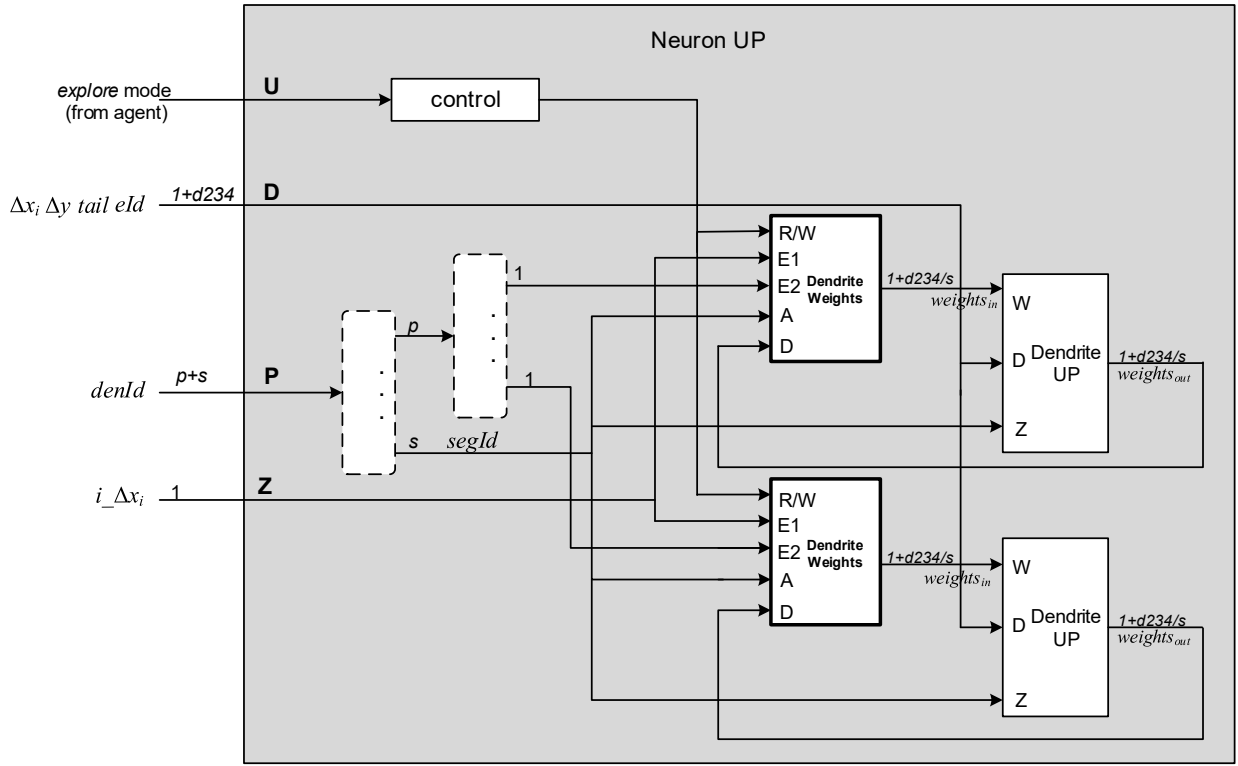

**Figure 24. Neuron synaptic update sub-block. Updates may occur during any cycle for which the mode is *explore*.**

## 7.3 Dendrites

Dendrite operation is split into two subfunctions just as neuron operation is. The inference subfunction is part of neuron inference and the dendrite update is part of neuron update..

### 7.3.1 Dendrite Inference

Dendrite inference was covered in Section 4.3. The schematic is repeated in Figure 25. The number of segments in a dendrite is independent of the other system dimensions, e.g. the sizes of distal and proximal inputs. A key determinant for the number of segments is the number of different contexts in which a given feature is expected to appear.

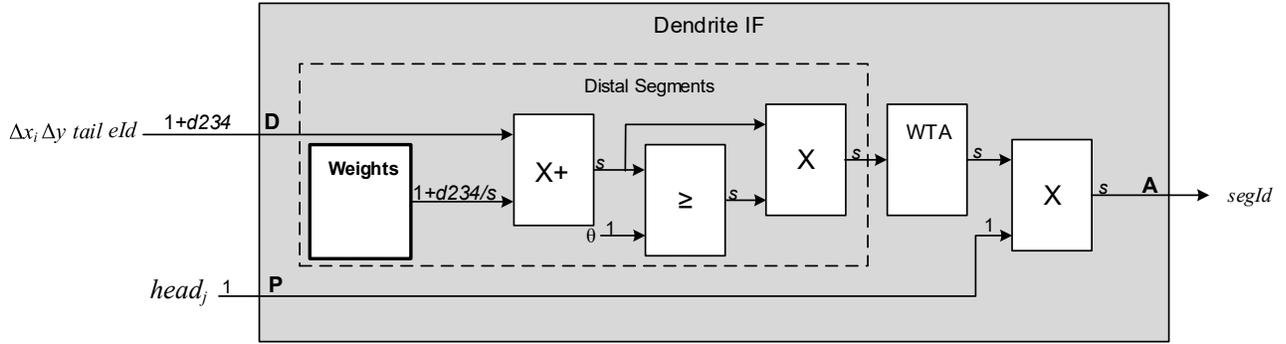

**Figure 25. Dendrite inference.**

### 7.3.2 Dendrite Synapse Update

The update schematic for the *capture* case is Figure 14. All three update cases are shown in Figure 26 . The cases are mutually exclusive, so their update values are easily merged in the *mrg* block. Then the upper and lower weight bounds ($w_{max}$ and 0) are enforced before the output is generated.

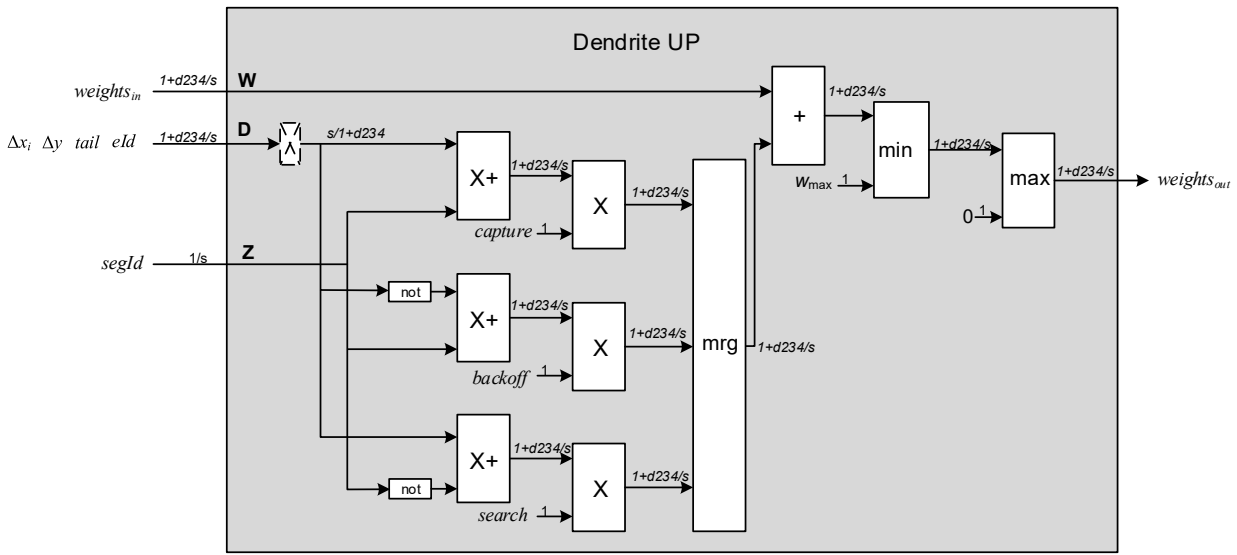

**Figure 26. Dendrite update schematic. All three update cases are shown.**

### 7.3.3 The Role of Similarity Clustering

Note that a single dendrite performs a function similar to a "column" in prior work [25]. In [25], it is shown that a "column" is capable of online unsupervised clustering. A sequence of input vectors is applied to a dendrite's distal inputs, and the dendrite organizes the vectors into clusters based on similarity, one cluster per dendritic segment. Using an SDP method as described above, cluster centroids are established and are encoded in the synaptic weights.

With good clustering, *similar* input vectors should yield the *same* output or action. In effect, similarity clustering is a form of lossy compression. With that in mind, observe that the research benchmark environments consist of randomly selected features placed at random locations. Randomness is the enemy of compression. Given the randomness in the problem specification, there are few similarities upon which quality clustering can be based. Consequently, in the first implementation of the research benchmark, every segment holds a single state vector, and every cluster has a single member, so no actual compression is performed.

Most real-life situations are not random, and the non-randomness often leads to good clustering opportunities. Follow-on research will focus on more realistic problems based on non-random environments where the full capabilities of SDP clustering can be better exploited.

### *7.4 Summary*

The complete architecture hierarchy for the PL block is illustrated in Figure 27. PL architecture hierarchy from top to bottom.

The function of PL is to learn environments during *exploration* and then to produce inferred outputs regarding the state when *orienting* or when *queried*.

*Exploration*

During exploration, the environment being explored, *eId_in*, is supplied externally – from the agent in this system, although it can come from anywhere outside the macrocolumn – and the *head* is the current feature (if there is one; otherwise the *head* is *null*). The agent roams the environment, and whenever *head* is non-null the necessary and sufficient condition for generating neuron spikes in the three minicolumns is satisfied and synapses are updated according to SDP rules to reflect the current state vector – an edge in the environment graph.

*Orientation*

The main objective of orientation is to determine the current *eId*. Initially, when the *eId* is unknown, the agent forces all bits in *eId_in* to be set, indicating that any of the environments is possible. The agent then roams the grid and as new features are reached, it applies the current state vector. The *i_eId* minicolumn responds with all the *eIds* in the current set that are consistent with the rest of the state vector. This eliminates all inconsistent *eIds* from the set of possibilities until only one potential *i_eId* remains (assuming the learned environments are not inherently ambiguous).

*Queries (Navigation)*

Assuming an unambiguous *eId* (orientation is finished) and that the system is located at some feature, a query consists of applying a target feature, and the *PL* responds with inferred $\Delta x$ and $\Delta y$ movements (*i_$\Delta x$*, and *i_$\Delta y$*) that will take the system from the current feature to the target feature, provided the associated edge in the environment graph has been learned. Otherwise, the query will return *null* values.

The specific queries that an agent makes and the way that it uses the results depend on the task the agent is attempting to achieve. In this document, the benchmark task is essentially trivial; implementing more elaborate tasks are beyond this document's scope and remain for follow-on research (see Section 9.1).

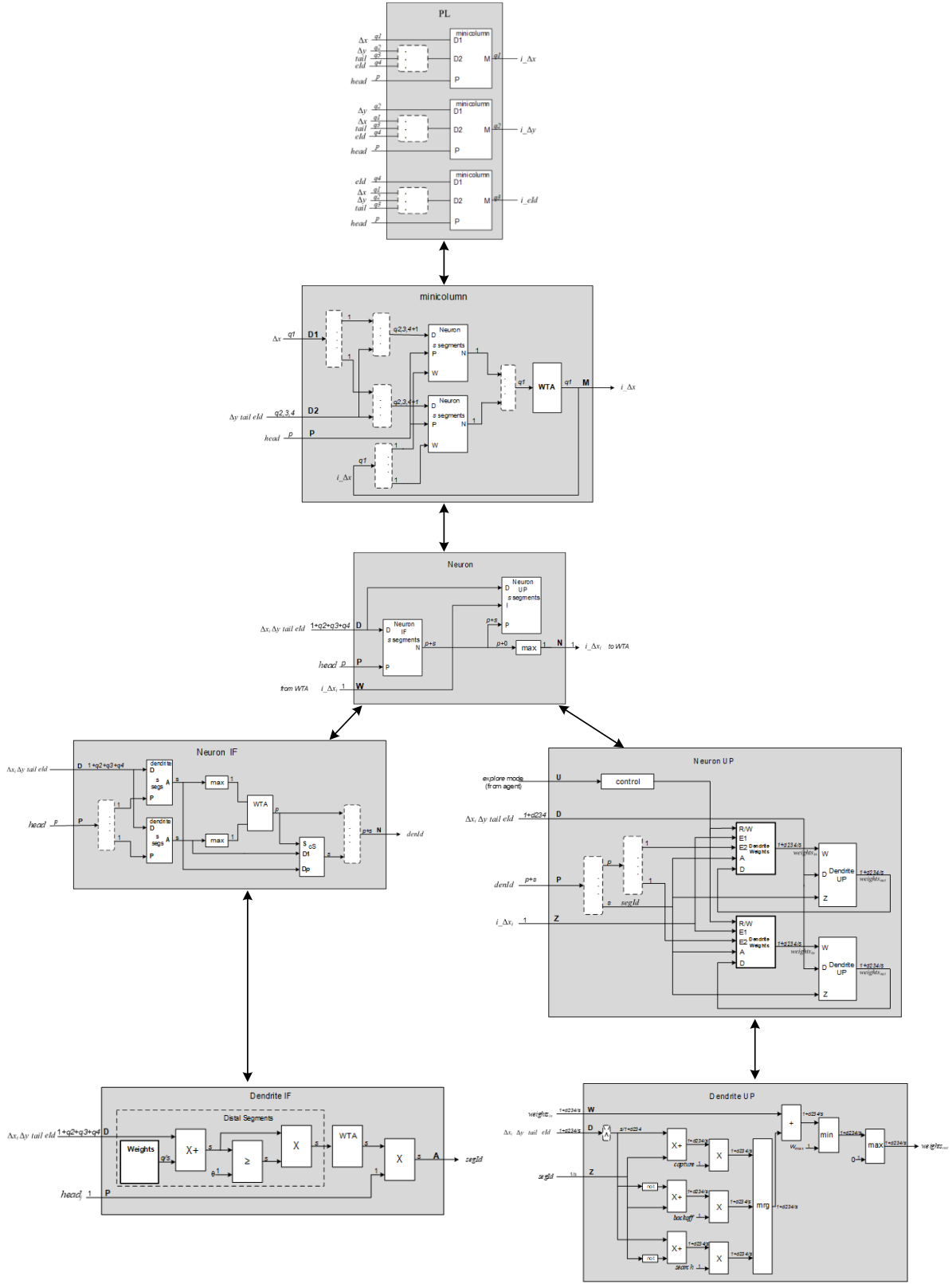

**Figure 27. PL architecture hierarchy from top to bottom.**

## 8. Simulations

First and foremost, benchmark simulations demonstrate that the proposed macrocolumn implementation works as described. Going beyond this demonstration, generating results for ranges of parameters and configurations, isn't especially enlightening because of the artificial nature of random environments and because a naive agent is used solely for the purpose of demonstrating a query mechanism that can support navigation, not for the purpose of achieving some specific objective.

### *8.1 Benchmark Details*

For a set of benchmark simulations, 40 different environments are laid out on a 30×30 grid. Each environment contains the same 10 features pseudo-randomly placed. The exploration path is designed to visit all 10 features four different times in a pseudo-random sequence.

Simulation consists of an exploration phase and an orientation/navigation phase. Simulations are further divided into episodes and steps. An *episode* is an exploration or navigation of a single environment and consists of a sequence of 100 steps, or gamma cycles. To expedite simulations, it is capable of moving an arbitrary number of grid cells in a single step. Because there are 40 environments, the simulation performs 40 exploration episodes followed by 40 navigation episodes.

After orientation, the agent initiates queries and movements by first querying with a random *target* to find $i\_\Delta x$ and $i\_\Delta y$. If there is a match in the PL, then the agent performs the inferred movements. Otherwise if the inferred movements are null, it tries another random *target*. Because the graph is connected, it will eventually succeed.

Because there is no similarity clustering and each segment is expected to hold a single state vector, the number of segments per dendrite is an important parameter. To get a handle on the number of required segments, the pseudo-random exploration paths for all 40 environments were first generated so that the eventual environment graphs could be determined. To account for random variations, the maximum number of segments over all three minicolumns was determined (skipping the details), and the number of segments in the simulator is set to that upper bound.

### *8.2 Simulation Results*

Because the number of segments allocated is sufficient to fit all the state vectors (by design), the main purpose of simulation is to demonstrate that the model functions as expected. Simulation parameters are given in Table 2.

**Table 2. Parameters for benchmark simulations**

| | |
|---|---|
| *threshold* θ | 8 |
| *initial weights* $w_b$ | 6 |
| *maximum weight* $w_{max}$ | 8 |
| *capture* | 1 |
| *backoff* | 4 |
| *search* | 0 |

A measured statistic is the number of steps it takes for orientation to complete; see Figure 28. Because there is significant variation in the environments, and movements depend on pseudo-random number generator seeds, three different seeds were simulated. Overall, most episodes infer the correct environment/feature in a few steps, although there are some outliers that take as long as 16 steps. Once oriented, accesses to *PL* during simulated navigation are always correct.

J. E. Smith                               31                               02/09/2023

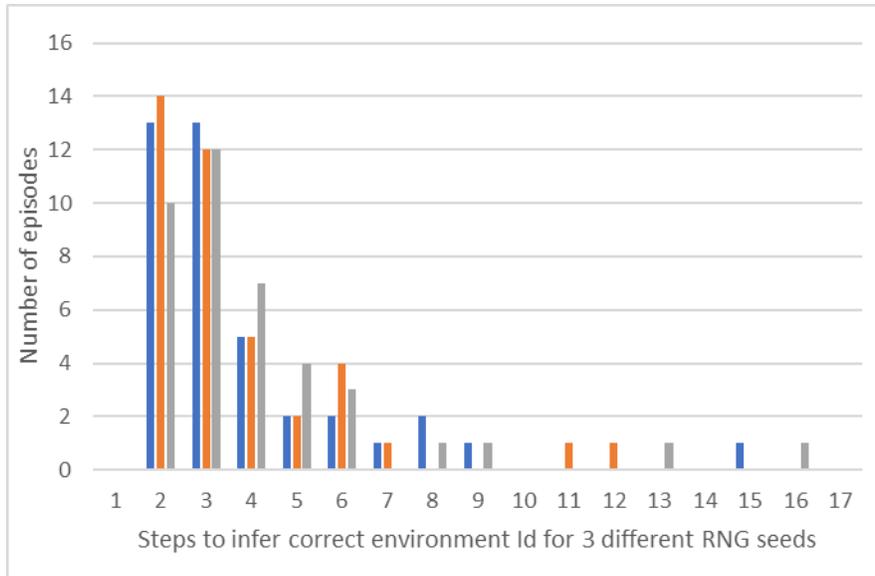

**Figure 28. The number of episodes required for orientation.**

If the number of segments in the design is not sufficient to hold all the state vectors, i.e. the segments are over-subscribed, then navigation failures may occur. A failure is detected, for example, if the macrocolumn gives displacements to a specific feature, but when the agent moves by the indicated displacements, the expected feature is not present. However, the macrocolumn is designed to handle such failures in a graceful way: when an error is detected the macrocolumn automatically re-sets itself to the state where all *eIds* are valid, re-orients itself, and continues on.

The plot in Figure 29 is for macrocolumns containing fewer segments than the calculated necessary number for all the state vectors. The statistic is the fraction of time the agent is correctly oriented after the initial orientation: its state vector indicates the correct *eId* and the feature for its current cell location is correct. For these simulations, the calculated lower bound on the sufficient number of segments per dendrite is 16. Figure 29 gives results for segment counts from 16 down to 4. Even with only 4 segments per dendrite, the agent is correctly oriented over 92% of the time.

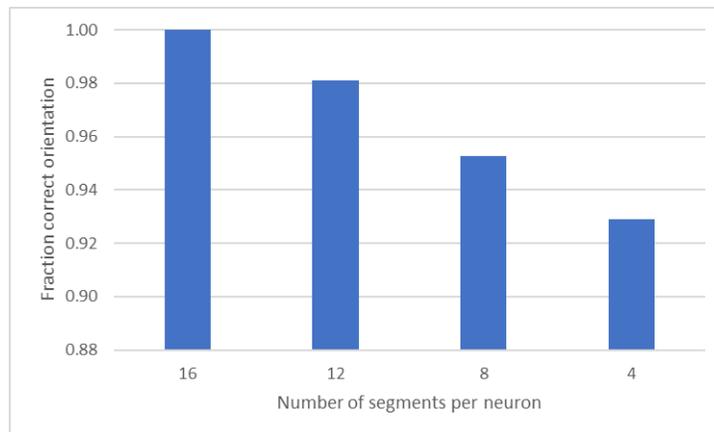

**Figure 29. Fraction of the steps that the agent is correctly oriented following its initial orientation as the number of segments per neuron is reduced.**

## 9. Concluding Remarks and Future Research

The basic macrocolumn learns and constructs directed graphs that describe 2-d spatial environments. The macrocolumn maintains a state vector that provides all the necessary information for learning a graph edge. It can then query the graph to support navigation. The model is based on spiking neurons and all the macrocolumn inputs and outputs are bit vectors.

What this document demonstrates is a bare-bones macrocolumn that supports the basic functions of exploration, orientation, and querying to support navigation. If the proposed macrocolumn accurately captures the fundamental way the neocortex works, then it can serve as a foundation for ever more elaborate brain-like cognitive systems. The bare-bones macrocolumn can also be characterized by its limitations:

      1) it does not include significant similarity clustering

      2) it does not support multi-macrocolumn regions or hierarchies

      3) it supports only a 2-d grid that mirrors a 2-d external environment

      4) it does not support invariance for scaling, orientation, and distortions

Furthermore, the agent is naive and is not constructed of model neurons. Following subsections discuss future research targeted at removing these limitations.

### *9.1 Agents*

In the research benchmark, the agent performs a minimal function: it moves about randomly from feature to feature. This is adequate for demonstrating the macrocolumn's ability to provide navigational information (inferred displacements), but not much else.

A topic for near-term research is the development of agents constructed of model neurons that strive to achieve specified objectives within a given environment. As a simple example, the presence of cheese at a grid point can be one of the learned features. Then, if the agent's objective is to find food and the mouse is placed at an arbitrary grid point in a previously learned environment, it moves around sensing the various features and then, having gotten its bearings, it heads along a near-optimum path (a sequence of edges in the environment graph) to the cheese. Similarly, if one of the features is water and another is a warm place to sleep, an agent can direct movement along a previously learned path. Here, the important enhancement to the model agent is the ability to learn near-optimal paths by using reinforcement learning (RL). The agent uses RL to learn the best next *target* for determining the next feature along a path to find food, water, or a warm grid point.

An additional enhancement is implementing an agent using minicolumn-like spiking neural networks that learn via a biologically plausible SDP version of RL. An RL agent capable of balancing an inverted pendulum is described in [6]. That implementation uses simple point neurons and does not distinguish distal and proximal synapses. By adding proximal synapses, the same RL agent can learn to achieve multiple objectives as shown in Figure 30.

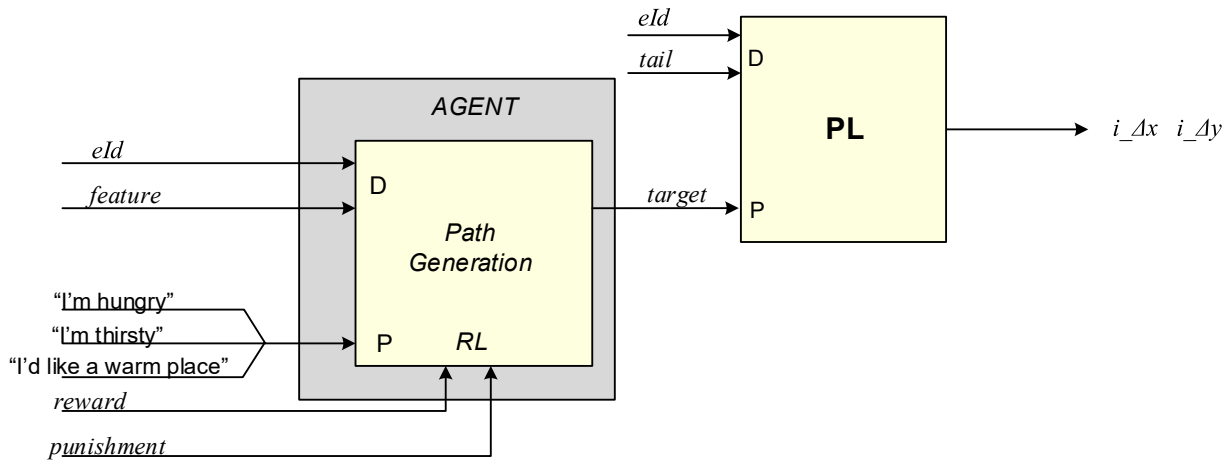

**Figure 30. An agent that uses reinforcement learning to follow optimal or near-optimal paths in order to achieve an objective.**

Say the metaphorical space aliens visit earth and examine a state-of-the-art CMOS microprocessor. Most of what they will see is SRAM – multi-level on-chip caches and predictors. And they may observe that the more SRAM, the better the performance (sometimes by huge amounts depending on working set sizes). They would then endeavor to discover how an SRAM works, motivated by the belief that it is the most important part of the computer. It is an essential part, to be sure, but the CPU is where the real computation takes place – it uses SRAM as a large data structure to support its operation.

The point is that even if we discover a complete and accurate model for the way the neocortex works, we may have only discovered the way "SRAM" works. It may be that the agent is primarily "CPU". And following this line of reasoning, the real key to understanding cognitive thought may reside in the agent, not the neocortex (although the agent may be partially based in the neocortex).

An RL agent as illustrated in Figure 30 is still relatively simple. What does an agent look like that drives language or mathematical processing based on macrocolumn frames? At a minimum this suggests that reverse-architecting research should incorporate the co-design of neocortical architecture and agent architecture.

### 9.2  Similarity Clustering

The macrocolumn developed in this document allocates a segment to each distal input pattern (context). Although the macrocolumn segments perform similarity clustering, strictly speaking, there is only one pattern per cluster. In a realistic application, the segment is trained to match multiple patterns forming a cluster, not just a single pattern. The synaptic weights in a segment define the cluster centroid. Consequently, all the similar patterns belonging to the same segment strongly match the segment's weights and cause a dendritic spike. Segments perform the same function as neurons in previous work [25][26], so implementing similarity clustering in dendritic segments should be straightforward.

Furthermore, for more realistic clustering benchmarks, environments with structured, non-random characteristics can be modeled. In other words, realistic problems where similarity clustering will offer significant benefits.

### 9.3  Regions

In this paper a single macrocolumn is considered. The next layer of abstraction is a *region* formed by interconnecting multiple macrocolumns. Figure 31 is a simple schematic for such a method.

This type of structure is considered by Lewis et al. [14]. For larger problem sizes, multiple macrocolumns operate in parallel, with each working on part of the larger problem – a simple example is a very large environment where macrocolumns track overlapping sections of the environment and where they collectively reach agreement regarding the overall environment identity via a form of distributed consensus ("voting" [14]).

Within a macrocolumn, the consensus mechanism uses a second set of distal inputs. Excitatory neurons have two biologically distinct sets of distal inputs: basal and apical. The distal inputs discussed in this paper are basal; distal inputs used for distributed consensus are apical [14].

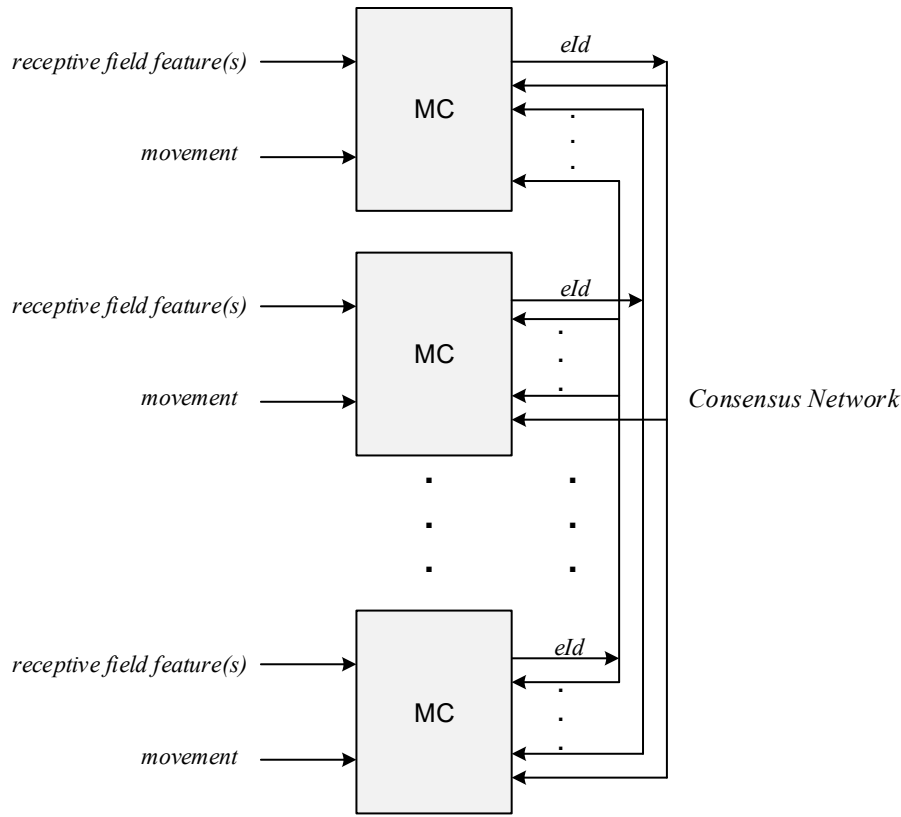

**Figure 31. A region consists of multiple macrocolumns (MC) that observe and move through overlapping receptive fields. A voting, or consensus, network implements distributed consensus to determine a region-wide environment identifier.**

### *9.4 Hierarchies*

Hierarchies are more easily discussed in terms of "objects" rather than "environments". For example, identifying an object by moving it around in one's hand is a function very similar to navigating an environment. Objects can be composed of features that are sub-objects having spatial relationships [11]. When reaching into a fishing pack for a fly box, the sharp corners, hinges, and clasp are identifiable objects by themselves based on their shapes. A macrocolumn (or macrocolumns) associates each of these objects with an *object Id*. At the next higher level of the hierarchy, the fly box as a whole is composed of these objects (now sub-objects) that have spatial relationships with each other, and at the higher level, the fly box has its own *object Id.* In Figure 15, the *eId* is a macrocolumn output that can be used for constructing such hierarchies.

## 9.5 Invariance

To achieve the brain-like ability to recognize objects, symbols, etc. there must be some method for implementing invariance with respect scales, rotations, and distortions. (It is already invariant with respect to spatial shifts). It seems that a good place to provide invariance is in the GD/path integration part of the macrocolumn. This suggests that various types of conformal mappings are built into the GD/path integration system. Going one step further, if the grid cell interconnection structure is wired-in, then that leaves path integration as the prime candidate for implementing invariance.

## 9.6 Non-sensorimotor Grids

Sensorimotor actions involve sensing and moving through a physical environment as studied here. For this kind of activity, modeling grid cells and path integration is relatively straightforward because both the external environment and the grid use 2-d coordinate systems.

However, "all thinking is movement" extends well beyond 2-d physical systems. For constructing macrocolumns suitable for higher level, more abstract cognitive tasks, the research becomes much more challenging. For such higher level cognitive tasks such as language processing, both a grid and a method for path integration will need to be discovered. Solving problems of this type and implementing them with macrocolumn architecture will require significant research breakthroughs. See [34] for a survey of research on constructing such cognitive maps.

## 10. References/Bibliography

[1] Abeles, Moshe. "Synfire chains." *Scholarpedia* 4, no. 7 (2009): 1441.

[2] Bi, Guo-qiang, and Mu-ming Poo. "Synaptic modifications in cultured hippocampal neurons: dependence on spike timing, synaptic strength, and postsynaptic cell type." *The Journal of neuroscience* 18, no. 24 (1998): 10464-10472.

[3] Butts, Daniel A., et al. "Temporal precision in the neural code and the timescales of natural vision." *Nature* 449, no. 7158 (2007): 92-95.

[4] Fries, Pascal, Danko Nikolić, and Wolf Singer. "The gamma cycle." T*rends in neurosciences* 30, no. 7 (2007): 309-316.

[5] Gardner, R.J., Hermansen, et al. Toroidal topology of population activity in grid cells. *Nature* **602,** 123–128 (2022).

[6] Gerstner, Wulfram, and J. Leo Van Hemmen. "How to describe neuronal activity: spikes, rates, or assemblies?" In *Advances in neural information processing systems*, (1993): 463-470.

[7] Gerstner, Wulfram, et al. "A neuronal learning rule for sub-millisecond temporal coding." *Nature* 383, no. 6595 (1996): 76-78.

[8] Hawkins, Jeff, and Sandra Blakeslee. *On intelligence*. Macmillan, 2004.

[9] Hawkins, Jeff, and Subutai Ahmad. "Why neurons have thousands of synapses, a theory of sequence memory in neocortex." *Frontiers in neural circuits* (2016): 23.

[10] Hawkins, Jeff, Subutai Ahmad, and Yuwei Cui. "A theory of how columns in the neocortex enable learning the structure of the world." *Frontiers in neural circuits* 11 (2017): 81.

[11] Hawkins, Jeff, et al. "A framework for intelligence and cortical function based on grid cells in the neocortex." *Frontiers in neural circuits* 12 (2019): 121.

[12] Hawkins, Jeff, A Thousand Brains: A New Theory of Intelligence, Basic Books, 2021.

[13] Hodgkin, Alan L., and Andrew F. Huxley, "A Quantitative Description of Membrane Current and Its Application to Conduction and Excitation in Nerve," *The Journal of physiology* 117.4 (**1952**): 500.

[14] Lewis, Marcus, et al. "Locations in the neocortex: a theory of sensorimotor object recognition using cortical grid cells." *Frontiers in neural circuits* 13 (2019): 22.

[15] Mainen, Zachary F., and Terrence J. Sejnowski. "Reliability of spike timing in neocortical neurons." *Science* 268, no. 5216 (1995): 1503-1506.

[16] Maass, Wolfgang, "Networks of spiking neurons: the third generation of neural network models." *Neural networks* 10.9 (1997): 1659-1671.

[17] McCulloch, Warren S., and Walter Pitts. "A logical calculus of the ideas immanent in nervous activity." *The bulletin of mathematical biophysics* 5, no. 4 (1943): 115-133.

[18] Morrison, Abigail, Markus Diesmann, and Wulfram Gerstner. "Phenomenological models of synaptic plasticity based on spike timing." *Biological cybernetics* 98, no. 6 (2008): 459-478.

[19] Moser, E. I., E. Kropff, and M. B. Moser. "Place cells, grid cells, and the brain's spatial representation system." *Annual Review of Neuroscience* 31 (2008): 69-89.

[20] Mountcastle, Vernon B. "The columnar organization of the neocortex." *Brain* 120, no. 4 (1997): 701-722.

[21] Natschläger, Thomas, and Berthold Ruf. "Spatial and temporal pattern analysis via spiking neurons." *Network: Computation in Neural Systems* 9, no. 3 (1998): 319-332.

[22] Saada, Ravit, et al. "Autaptic excitation elicits persistent activity and a plateau potential in a neuron of known behavioral function." *Current Biology* 19, no. 6 (2009): 479-484.

[23] Sanders, Honi, et al. "Grid cells and place cells: an integrated view of their navigational and memory function." Trends in neurosciences 38, no. 12 (2015): 763-775.

[24] Smith, James E. "(Newtonian) Space-Time Algebra." *arXiv preprint arXiv:2001.04242* (2019).

[25] Smith, James E. "A Neuromorphic Paradigm for Online Unsupervised Clustering." *arXiv preprint arXiv:2005.04170* (2020).

[26] Smith, James E. "A Temporal Neural Network Architecture for Online Learning." *arXiv preprint arXiv:2011.13844v2* (2020).

[27] Smith, James E. "Implementing Online Reinforcement Learning with Temporal Neural Networks." *arXiv preprint arXiv:2204.05437* (2022).

[28] Smith, James E. "Efficient digital neurons for large scale cortical architectures." *2014 ACM/IEEE 41st Annual International Symposium on Computer Architecture (ISCA)*, pp. 229-240, IEEE 2014

[29] Smith, James E. "Space-time computing with temporal neural networks." *Synthesis Lectures on Computer Architecture* 12, no. 2 (2017): i-215.

[30] Thorpe, Simon J. "Spike arrival times: A highly efficient coding scheme for neural networks." *Parallel processing in neural systems* (1990): 91-94.

[31] Thorpe, Simon J., and Michel Imbert. "Biological constraints on connectionist modelling." *Connectionism in perspective* (1989): 63-92.

[32] Tuckwell, Henry C. "Synaptic transmission in a model for stochastic neural activity." *Journal of theoretical biology* 77.1 (1979): 65-81.

[33] Watkins, Christopher JCH, and Peter Dayan. "Q-learning." *Machine learning* 8, no. 3 (1992): 279-292.

[34] Whittington, James CR, David McCaffary, Jacob JW Bakermans, and Timothy EJ Behrens. "How to build a cognitive map." *Nature Neuroscience* (2022): 1-16.

[35] Zador, Anthony M. "A critique of pure learning and what artificial neural networks can learn from animal brains." *Nature communications* 10, no. 1 (2019): 1-7.

## Appendix 1: Einschematics

"Einschematics" draw their inspiration from "einsums", a concise notation for expressing summations, popularized by machine learning frameworks such as PyTorch.

They are a concise and unambiguous schematic notation for spiking neural networks. Einschematics operate on scalars, vectors, and 2-d arrays. Figure 32 contains operators that operate only on lines, so the number of block inputs and outputs are always the same. All indexing begins at 1.

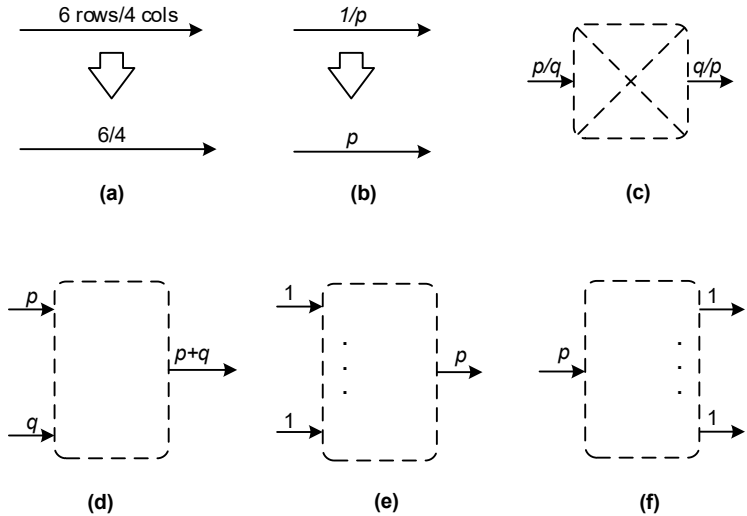

**Figure 32. Einschematic blocks that operate on lines are drawn with dashed lines a) In general, lines carry 2-d arrays, denoted as *#rows* / *#columns*. b) If the number of rows is 1, i.e. 1 / *#columns*, this is simplified to *#columns*. A scalar has one row and one column, 1/1, and is denoted as 1. c) the transpose of a 2-d array. d) a concatenation operator that merges a bundle of *p* lines with a bundle of *q* lines to form a bundle of *p+q* lines. e) The special case where *p* scalars are merged to form a size *p* vector. f) A size *p* vector is split into *p* scalars (size 1 vectors).**

Examples of einschematic functional blocks are in Figure 33. The operation to be performed is a function of both the given operator and the dimensions of the inputs and the outputs. All lines leaving a block are labeled with their dimensions, and these labels are an essential part of a schematic. For non-commutative operations, such as division, subtraction, and matrix multiplication, the first operand is the top input.

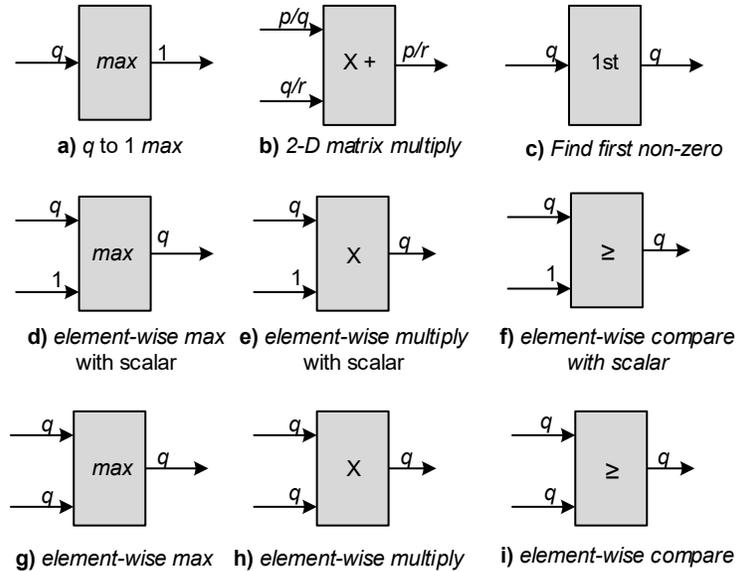

Figure 33. Examples of einschematic functional blocks. Three versions of the *max* operation are shown in a), d), and g). Their given operator (*max*) is identical for all 3. However, the numbers of inputs and outputs imply the specific version of *max*. The *max* in a) reduces the $q$ input lines to their maximum value, output as a scalar on a single line. The *max* in d) performs the max of a size $q$ input vector and a scalar, producing an output of $q$ lines. Finally, in g) an elementwise vector *max* is performed. A 2-D matrix multiplication consisting of a × followed by + is illustrated in b). The blocks e) and h) are different versions of multiplications with the type of operation being deduced from the input and output line sizes. Operations involving predicates, as in f) and i) always produce bit vectors as outputs. The find first non-zero operator c) is essentially a generalized priority encoder. The output is a vector whose $k$th element equals the value of the $k$th element of the input vector, where $k$ is the smallest index of a non-zero element in the input. If there is no non-zero element in the input, then the output is set to all zeros. A similar function (not shown) passes a single pseudo-randomly selected non-zero input element $k$ to output element $k$.

Examples of functional blocks are given in Figure 34. The notes "integers" and "bits" are not a requirement, but they could be added optionally to specialize the functional block implementation according to its inputs' properties. Similarly, notations of "sparse" or "dense" may be added for specializing functional block implementations.

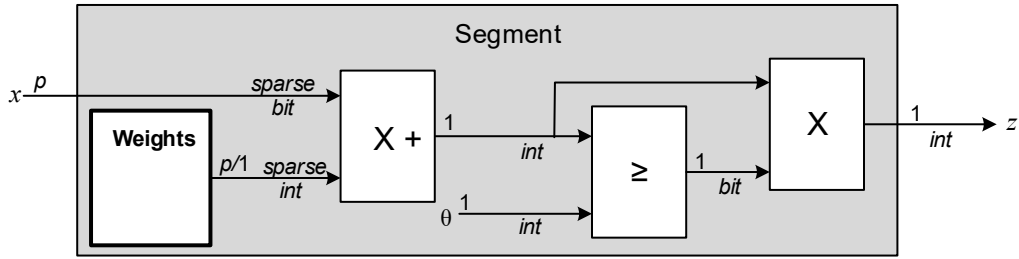

a) **Dendritic segment consisting of a single point integrator. Annotations regarding type (bit/int) and sparsity are not part of the functional specification. They provide implementation information, i.e., the implementation of the function may be optimized to best suit the properties of its operands.**

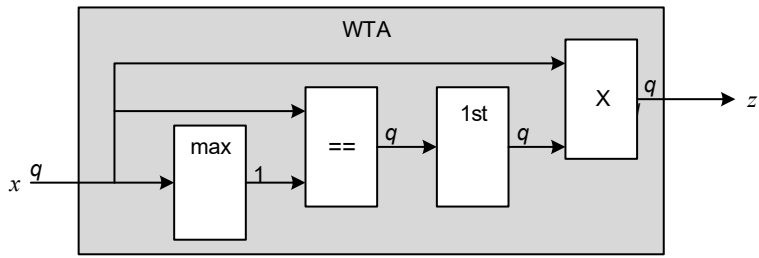

b) **WTA block. Ties are selected by using the least index of the tying inputs.**

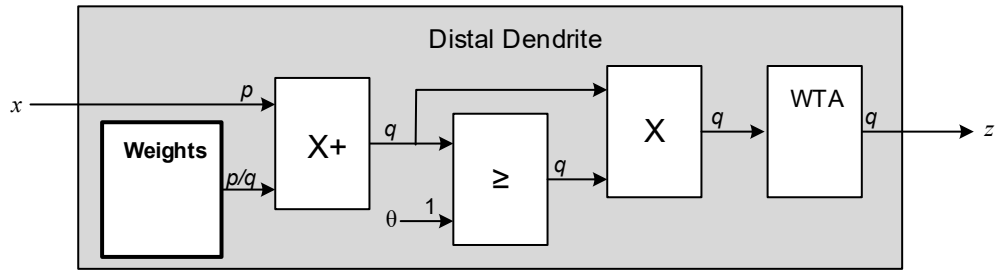

c) **A dendrite consisting of multiple dendritic segments.**

**Figure 34. Einschematic examples: three functional blocks.**

**Appendix 2: Relationship to Prior Work: Temporal Communication and Processing**

In a *temporal* neural network model [25]-[29] values are encoded and communicated as temporal volleys (vectors) where vector elements correspond to relative spike times. Infinity is used when there is no spike (see Figure 35a). A temporal encoding has a given precision; in Figure 35a where spike times vary from 0 to 7 there are 3 bits of temporal precision.

The temporal precision that can be maintained in a biological system has a significant bearing on its computational capabilities and the complexity of the computational processes. Generally speaking, with respect to a hardware implementation, the lower the precision, the lower the implementation complexity. If this is taken to an extreme, one bit of temporal precision should yield the simplest system. A single bit of temporal precision yields a *binarized* communication model (Figure 35b), where spikes are represented as binary 1's and non-spikes as 0's.

In exchange for the simplicity of a single bit of temporal precision, an important issue is how much computational capability is lost in the process. This is addressed in following paragraphs.

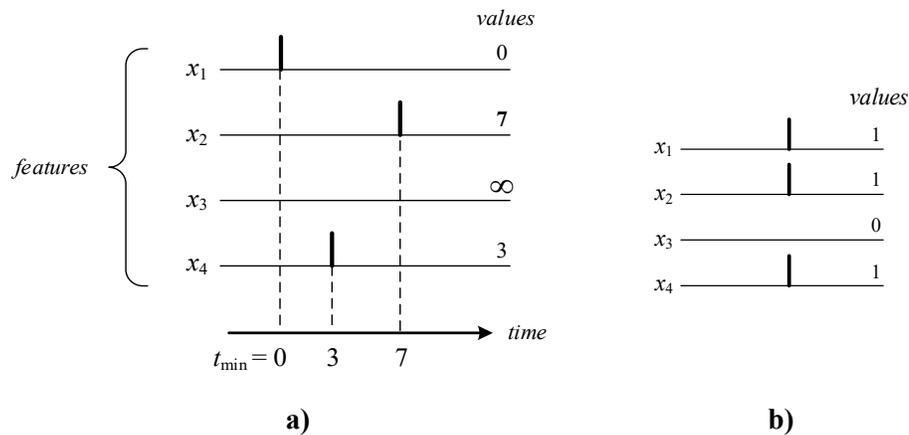

**Figure 35. Spike volleys. a) Temporal volleys: a spike indicates the presence of a feature, and the time of a spike indicates the feature's relative strength, with 0 being the strongest. b) Binarized volleys: features are either present (1) or absent (0).**

*Temporal Neuron Implementations*

With temporal modeling, point neurons are typically modeled with some form of Spike Response Model (SRM) proposed in 1994 by Gerstner and van Hemmen [6]. A simple form (SRM0) is illustrated in Figure 36. In the SRM0 model, input $x_i$ connects to the neuron body via synapse $i$ having weight $w_i$. 1) if there is a spike on $x_i$, then the value (time) of the associated synaptic weight $w_i$ selects a pre-defined *response function*; 2) in the neuron body, the synaptic response functions are *summed* (*integrated*), yielding a net *body potential*; 3) when, and if, the body potential reaches a *threshold* value θ, a spike on output $z$ is produced at that time.

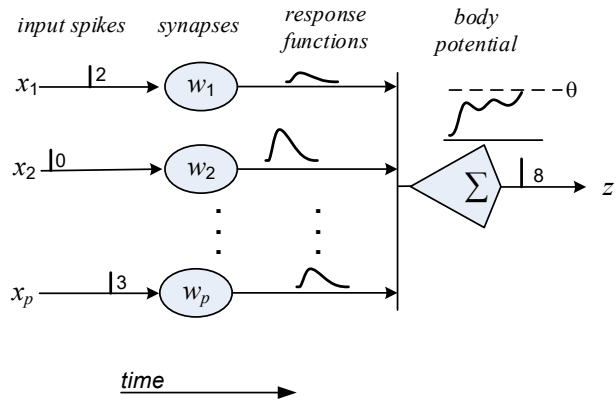

**Figure 36. SRM0 neuron model.**

A number of different response functions have been used by researchers over the years (Figure 37). In the course of prior research, the author has progressed through each of these. Early on, a biexponential response function was used (Figure 37a); it is the more biologically accurate. Then, the biexponential was approximated with the piecewise linear version (Figure 37b), proposed by Maass [16]. It is this version that is used in the book [29]. In subsequent research, this was replaced with the ramp-no-leak version in [25]-[27]. The reasoning behind this simplification is that the negative slope due to neuron body leakage is only a reset mechanism and is not a part of active computation.

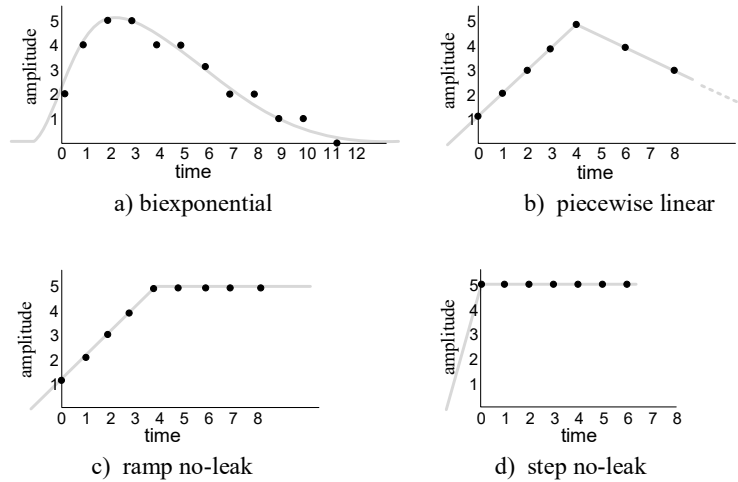

a) biexponential    b) piecewise linear

c) ramp no-leak    d) step no-leak

**Figure 37 Four response functions.**

*Binary Neurons*

Assume neural networks that take binary vectors as inputs and consider an SRM0 neuron implementation with a ramp no-leak response function.

In practice, synaptic weights typically stabilize in a bimodal distribution (close to 0 or $w_{max}$). This means that in most cases the response functions for non-zero synapses are the same. For weight $w_{max}$, assume a ramp response function $\rho(t) = 1 + t$ ; $t \geq 0$. With binarized inputs, all ramp responses begin at $t = 0$. Assuming $m$ spikes are input to maximum weight synapses, the body potential $v$ is the sum of $m$ response functions $v(t,m) = m + m*t$. An output spike is produced at the time the threshold $\theta$ is reached; for $m$ matches, define this to be $t_m$, so the condition for outputting a spike is: $m + m* t_m \geq \theta$. Because times are

J. E. Smith 43 02/09/2023

integers in the model: $t_m = \lceil \theta/m \rceil - 1$. This is an inverse relationship: the spike time is inversely related to the number of matches (i.e., the total potential).

We can take advantage of this relationship by using the total potential as a neuron's output and spare the effort of computing the inverse. Furthermore, when winner take all (WTA) inhibition is performed over a set of parallel neurons (a "column" in the cited prior work), spike time ties are broken by taking the highest body potential amongst the tying neurons. This means that the body potential has to be passed to WTA, anyway. Consequently, it may be computationally simpler to compute and output the body potential directly and let WTA operate over body potentials. The WTA output can then be binarized, or, alternatively the body potential can be passed on to downstream processing. This is consistent with the more conventional temporal neuron implementation, and, in effect, models a temporal system, albeit at a higher level of abstraction where the temporality is abstracted away.

If a neuron takes binarized inputs and outputs body potential, a bimodal weight distribution means that virtually any well-behaved SRM0 response function (e.g. all those in Figure 37) will work. This being the case, one can use a step no-leak response function (Figure 37d). And this, in turn, is equivalent to taking a simple dot product of a binary input vector and the weights.

In a binarized system where the temporal precision for *inter*-neuron communication is a single bit, *intra*-neuron computation can employ significantly higher precision; six to eight bits, say. That is, a computational block containing neurons and WTA inhibition can take binary input vectors and produce binary output vectors, but use higher precision integer vectors internally -- when forming the dot product of binary inputs and weights, for example.

Finally, it is important to note that with binary inputs both the temporal method using a ramp no-leak response function and the binary dot product computation method are equally accurate when combined with WTA inhibition where ties are broken based on body potential. The choice of which to use depends on the implementation technology and engineering tradeoffs.

*Plausibility*

There is an important plausibility issue regarding the precision of inter-neuron temporal communication. That is: how much precision can be maintained as spikes pass from neuron to neuron?

If we assume that inhibitory gamma cycles [4] separate each neuron-to-neuron communication, then the window for temporal encoding is about 5-10 msec., because the gamma cycle is 50-100 Hz and half the cycle (at least) is consumed by the inhibitory phase. Experiments such as Mainen and Sejnowski [15] show that if an excitatory neuron is repeatedly stimulated with the same input pattern, then the output spike times repeat to within 1 msec. This implies individual neurons are capable of 2-3 bits of temporal precision, so it may be possible to maintain 2-3 bits of accuracy for intra-neuron processing.

Next consider the inter-neuron level. A spike with a resolution of 1 msec may be emitted from a neuron's body, but then it must travel down the axon, interact with a synapse to open a conductive gate and, along with other spikes, generate a dendritic pulse that reaches the downstream neuron's body and eventually results in an output spike. Is it plausible that 1 msec temporal resolution can be maintained throughout this sequence of events that crosses a gamma cycle boundary? It seems more plausible that only 1 bit of temporal precision is sustainable for inter-neuron communication. This implies the full encoding window of 5-10 msec is used for communicating a single bit. In a temporal system, this 1 bit of precision is equivalent to a binarized inter-neuron communication system.